\documentclass[acmsmall]{acmart}
\usepackage{graphicx}
\usepackage{hyperref}
\PassOptionsToPackage{numbers,sort&compress}{natbib} 
\usepackage{natbib}
\usepackage{mdframed}
\usepackage{multirow}
\usepackage{makecell}
\usepackage{booktabs} 
\usepackage{array}
\usepackage{pdflscape} 
\usepackage{adjustbox}
\usepackage{hyperref}
\hypersetup{hidelinks,
	colorlinks=true,
	allcolors=blue,
	pdfstartview=Fit,
	breaklinks=true}

\setcopyright{acmlicensed}
\copyrightyear{2025}
\acmYear{2025}
\acmDOI{XXXXXXX.XXXXXXX}

\acmVolume{1}
\acmNumber{1}
\acmArticle{1}
\acmMonth{1}

\begin{document}
\title{Graph2text or Graph2token: A Perspective of Large Language Models for Graph Learning}

\author{Shuo Yu}
\affiliation{
\department{School of Computer Science and Technology}
  \institution{Dalian University of Technology}
  \country{China}
}
\email{shuo.yu@ieee.org}

\author{Yingbo Wang}
\email{yingbo.wang.dlut@outlook.com}
\author{Ruolin Li}
\email{liruolin@mail.dlut.edu.cn}
\affiliation{
\department{School of Software}
  \institution{Dalian University of Technology}
  \country{China}
}

\author{Guchun Liu}
\email{liuguchun@mail.dlut.edu.cn}
\author{Yanming Shen}
\email{shen@dlut.edu.cn}
\affiliation{
\department{School of Computer Science and Technology}
  \institution{Dalian University of Technology}
  \country{China}
}

\author{Shaoxiong Ji}
\affiliation{
\department{Department of Computer Science}
\institution{Technical University of Darmstadt}
\country{Germany}
}
\email{shaoxiong.ji@tu-darmstadt.de}

\author{Bowen Li}
\email{bowen.li@rmit.edu.au}
\author{Fengling Han}
\email{fengling.han@rmit.edu.au}
\author{Xiuzhen Zhang}
\email{xiuzhen.zhang@rmit.edu.au}
\author{Feng Xia}
\authornote{Corresponding Author}
\email{f.xia@ieee.org}
\affiliation{
\department{School of Computing Technologies}
\institution{RMIT University}
\country{Australia}
}

\renewcommand{\shortauthors}{Yu et al.}

\begin{abstract}

Graphs are data structures used to represent irregular networks and are prevalent in numerous real-world applications.
Previous methods directly model graph structures and achieve significant success.
However, these methods encounter bottlenecks due to the inherent irregularity of graphs.
An innovative solution is converting graphs into textual representations, thereby harnessing the powerful capabilities of Large Language Models (LLMs) to process and comprehend graphs.
In this paper, we present a comprehensive review of methodologies for applying LLMs to graphs, termed LLM4graph.
The core of LLM4graph lies in transforming graphs into texts for LLMs to understand and analyze.
Thus, we propose a novel taxonomy of LLM4graph methods in the view of the transformation.
Specifically, existing methods can be divided into two paradigms: Graph2text and Graph2token, which transform graphs into texts or tokens as the input of LLMs, respectively.
We point out four challenges during the transformation to systematically present existing methods in a problem-oriented perspective.
For practical concerns, we provide a guideline for researchers on selecting appropriate models and LLMs for different graphs and hardware constraints.
We also identify five future research directions for LLM4graph.
\end{abstract}




\maketitle

\section{Introduction}
\label{sec:introduction}
A graph is a data structure used to represent objects as nodes and their relationships as edges \cite{sahu2020ubiquity}.
In contrast to images or texts, graphs are naturally irregular since different nodes have distinct node neighbors.
Such special irregularity makes graphs to be widely used to represent complex data systems across various domains \cite{9416834,10026151}, such as social networks \cite{myers2014information}, molecules \cite{yi2022graph}, knowledge graphs \cite{tiddi2022knowledge,10411654}, recommender systems \cite{ying2018graph,XiaAccess2013}, and citation networks \cite{shi2015vegas,YUPave201838}.
Researchers have developed numerous deep learning models for learning representation on graphs for downstream tasks.

Directly modeling graphs for representation learning encounters bottlenecks because of the inherent irregularity of graphs.
Graph Neural Networks (GNNs) \cite{kipf2016semi, velivckovic2017graph} directly model irregular node neighborhoods, but they struggle in over-smoothing \cite{rusch2023survey}, over-squashing \cite{di2023over}, and under-reaching \cite{alon2021on}. 
Graph Transformers (GTs) \cite{ying2021transformers,shehzad2024graph} further model potential interactions between distant nodes, but they encounter over-globalizing \cite{xing2024less} and high complexity \cite{shehzad2024graph}.
Furthermore, the irregularity induces special properties for graphs (e.g., permutation invariance), compelling models to specialize in sophisticated graph structures \cite{GNNBook2022}.
Although GNNs and Graph Transformers adeptly capture structural signals, they often struggle with preserving the richer semantic relationships (e.g., who is friends with whom and why), leading to a semantic gap. 

These limitations highlight the need for more robust, semantically-aware methods to model graphs as texts. 
Pioneering studies analogize nodes to words and then adapt a language model to optimize node co-occurrence \cite{perozzi2014deepwalk, grover2016node2vec}.
These seminal works sparked widespread interest and achieved considerable results, showing the feasibility and potential of modeling graphs as texts.
Recently, the advent of Large Language Models (LLMs) has revolutionized this line of research through their world-wise knowledge and dominant text-processing capabilities \cite{sun2024head, huang2023finbert}.
LLMs go beyond graph structures and profoundly comprehend graphs as human understandable texts.
Consequently, LLMs greatly improve the performance of graph representation learning, signifying a groundbreaking advancement in modeling graphs as text.

Building on these early successes in treating graphs as text, we now explore how modern LLMs specifically accommodate graph structures. LLMs show excellent compatibility with graph data.
Specifically, LLMs receive inputs with different grains: texts and tokens.
Some graphs contain rich textual information (e.g., citation networks and knowledge graphs) \cite{buneman2021data, hogan2021knowledge}, and graphs are always constructed in a specific context (e.g., user interactions in social networks) \cite{wilson2009user, kleinberg1999web}.
As a result, graph structures can be seamlessly represented as texts for LLMs to comprehend.
Furthermore, the hierarchy of graph structures (nodes, edges, and sub-graphs) parallels to that of tokens (characters, words, and sentences) \cite{he2024unigraph,feng2022kalm,zhou2020graph,tang2024graphgpt}.
Consequently, graph structures can be flexibly transferred into tokens and their vector representation.

LLMs improve both the performance and the generalization of graph representation learning.
To input graphs to LLMs, graph structures must be first transformed into textual representation.
Thus, these textual representations inherently contain structural information of graphs.
Consequently, LLMs thoroughly understand graphs in the views of both structures and semantics, significantly improving performance in downstream tasks.
The irregularity of graphs induces inconsistent node identities in different graphs, resulting in a generalization barrier for representation learning \cite{zhou2020graph}. 
Nevertheless, LLMs can naturally generalize across graphs in different domains through their abundant prepared knowledge.

\begin{figure}[htbp]
    \centering
    \includegraphics[width=0.9\linewidth]{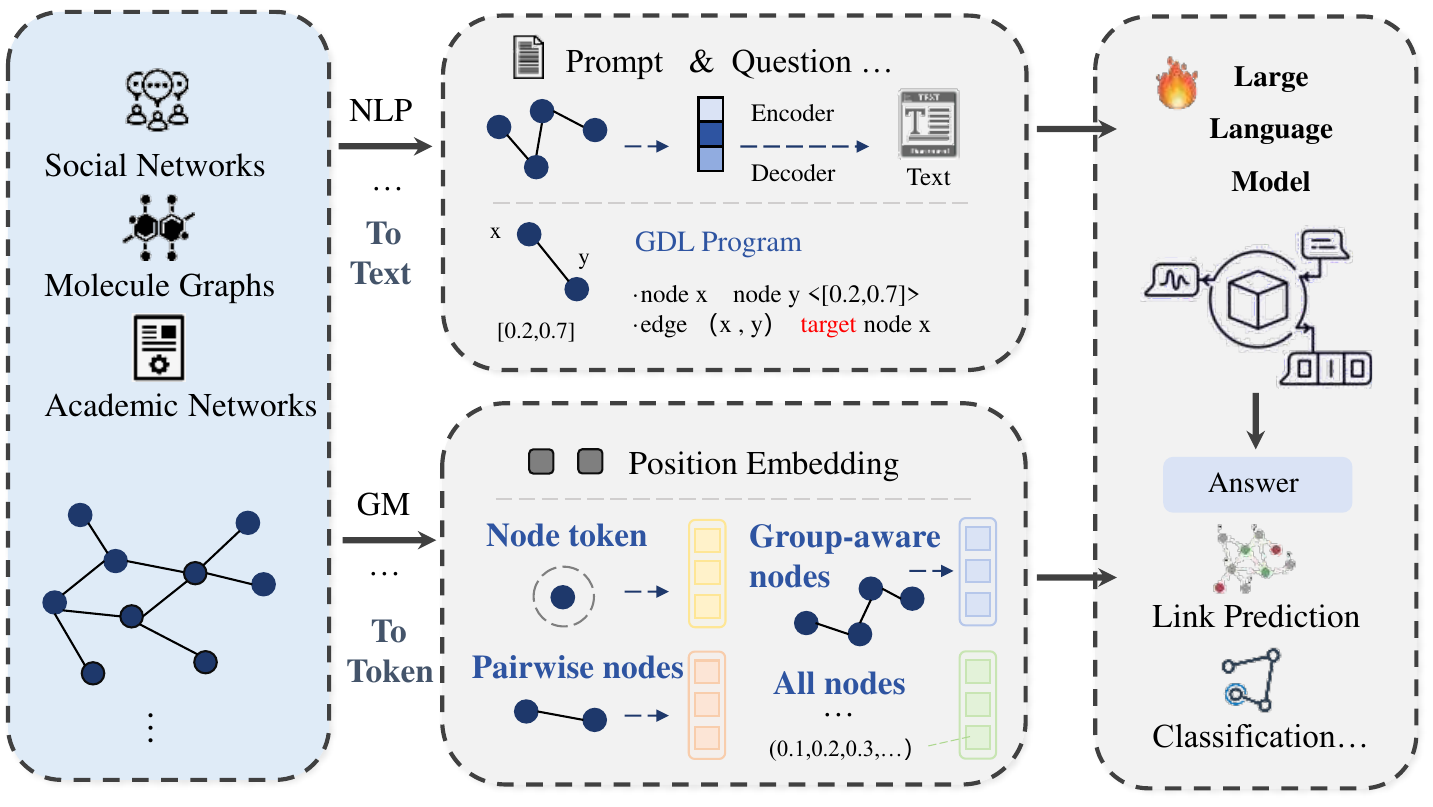}
    \caption{The illustration of LLM4graph. LLMs understand and process various types of graph data through Graph2text or Graph2token.}
    \label{fig:intro}
\end{figure}

In this paper, we systematically review how LLMs are applied for graphs, termed LLM4graph.
Since LLM4graph is rooted in the fact that graphs can be modeled as texts, the core of LLM4graph lies in how to transform graphs into texts.
Thus, this review proposes a novel taxonomy of existing LLM4graph methodologies from the perspective of the transformation from graph to texts.
In this paper, the general term `text' virtually refers to two aspects of textual information: texts (the raw human-readable language) and tokens (the fundamental units used by language models to process and understand textual data.). 
\footnote{The term `text' in this paper refers to the general meaning as a kind of data modality or a specific kind of textual representation. The former may present as `modeling/transforming graphs as/to texts', and the latter may present as `Graph2text'. These two meanings are used alternatively.}
As shown in Fig. \ref{fig:intro}, existing methods can be divided into two paradigms: Graph2text and Graph2token.
Specifically, Graph2text transforms graph structures into readable text, allowing LLMs to understand and even directly apply to downstream tasks.
Concurrently, Graph2token transforms graphs into tokens or their embedding as the input of LLMs through pre-trained or graph-based models.
While Graph2text adopts human-readable texts to guarantee interpretability, Graph2token utilizes fine-grained representation to fully fuse structural and semantic information.
We subclassify these two kinds of methods in a problem-oriented perspective.
Concretely, we point out four challenges when transforming graphs into texts: alignment, position, multi-level semantics, and contexts.
Notably, we also review some potential methods that can be applied to LLM4graph.

Our review differs significantly from existing surveys, which generally concentrate on the roles of LLMs in specific model architectures  \cite{liu2023towards, li2023survey, jin2023large, ren2024survey, guo2024learning} or downstream tasks \cite{zhang2023graph2, agrawal2023can, wei2024towards, fan2024graph}. Although these works illuminate where LLMs are applied, they often do not explain why and how LLMs fundamentally interpret and process graph structures.
In contrast, we examine the underlying reasons that make LLMs well-suited for graph data and show how existing methods address the challenges of modeling graphs as textual representations. Afterwards, we subsequently provide a practical guideline of LLM4graph for researchers.
We summarize how to choose LLMs and adopt corresponding tricks (prompts and fine-tuning) and select appropriate encoders for different graphs.
We also mark five future directions that researchers should consider in practice.

Our contributions are summarized as follows:
\begin{itemize}
    \item
    We explain why and how LLMs can fundamentally be applied to graphs from a problem-oriented perspective.
    We systematically survey how existing methods bridge gaps between two modalities when they model graphs as textual representations.
    \item
    We propose a new taxonomy for LLM4graph based on the transformation from graphs to textual representation.
    Specifically, LLM4graph can be divided into two branches of methods: Graph2text and Graph2token.
    Beyond existing work, we also survey potential methods that can be applied to LLM4graph.
    \item
    We offer recommendations on choosing suitable encoders for different graph types, along with technical tips (e.g., prompts, fine-tuning) to boost performance and interpretability. We also compile open-source resources and note computational considerations.
    \item
    We outline five open problems for the LLM4graph community, aiming to inspire novel research and broader adoption of LLM-driven graph learning techniques.
\end{itemize}

\raggedbottom
\begin{figure}[htbp]
    \centering
    \includegraphics[width=1\linewidth]{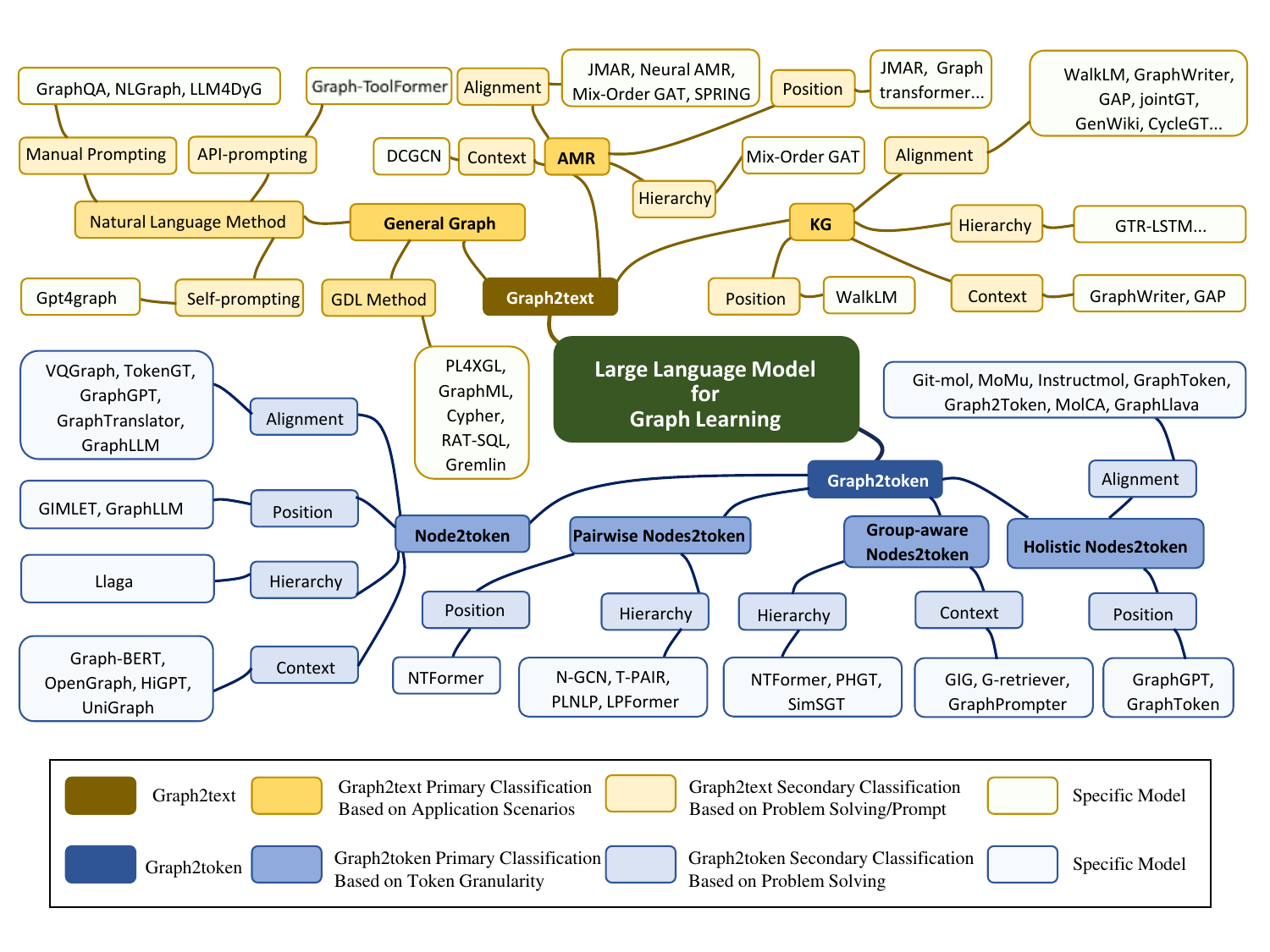}
    \caption{A taxonomy of research on large language models for graph learning.}
    \label{fig:example2}
\end{figure}

\section{Preliminaries}
\label{sec:Preliminaries}
In this section, we introduce the foundational concepts relevant to graphs and LLMs.
We give a formal definition of graphs in Section \ref{graph}, along with two common graph types.
In Section \ref{LLM}, we define LLMs and discuss the concepts of text, token, prompt, and context received by LLMs, and the general process of LLMs to handle two different input formats. 
Finally, Section \ref{encoder} describes the common encoders that are used to transform graph data into text or tokens compatible with LLMs.

\subsection{Graphs: Definitions and Common Types}
\label{graph}
Graphs are a typical type of non-Euclidean structured data. We state that the graphs are defined as: $G = \langle V, E \rangle$, where $V$ is the set of nodes in graph $G$, denoted as $\{v_1, \ldots, v_n\}$, and $E$ is the set of edges in graph $G$, denoted as $\{e_1, \ldots, e_m\}$. In the rest of this section, we introduce Abstract meaning representation (AMR) graphs and Knowledge graphs (KGs) which are two important graph structures in Graph2text.

AMR graph is essentially a single-rooted directed acyclic graph (DAG), which can clearly capture the contextual semantics of the article through its structure. Formally, an AMR graph can be defined as follows: $G_{AMR}=(V,E,r)$, where $V=\{v1,v2,\ldots,vn\}$ is a set of nodes, $E=\{(v_{i},v_{j},l_{ij}) \mid v_{i},v_{j}\in V,l_{ij}\in R\}$ is the set of labeled edges, $r\in V$ is the root node of the graph. Each node and edge may carry labels to further describe their meanings. Common labels include:  $root$ (representing the core concept of the entire expression), $ARGX$ ($ARG$ often indicates the agent of an action, and $X$ marks the argument role number), $opX$ ($op$ represents operations or functions, $X$ represents the parameter), and contextual relationship labels. For example, in the AMR diagram in Figure \ref{FIG4}, the node $v_2$ ("fights") is the root, $v_1$ ("Harry") is the $:ARG0$, and $v_3$("bravely") is the $:manner$.  

KGs describe entities, relationships, and attributes to represent and store real-world knowledge in a structured manner, where nodes represent entities and edges represent relationships. Formally, a KG can be defined as a set of triple: \( G_{KG}=\{(e,r,e') \mid e,e' \in \epsilon, r \in R\} \), where \(\epsilon\) and \(R\) denote the entity set and relation set, respectively.
\subsection{Large Language Models: Definitions, Inputs, and Processing}
\label{LLM}

LLMs are advanced computational frameworks trained on extensive text corpora to learn the statistical relationships between words and phrases. They are capable of understanding and generating coherent text, leveraging both syntax and semantics. In this survey, we focus on generative LLMs (e.g., Chatgpt and Llama \cite{touvron2023llama}), which produce new content based on prompts and can be further adapted or fine-tuned for specialized tasks.

LLMs typically accept two broad forms of textual representation as inputs:
\begin{itemize}
    \item \textbf{Text} refers to human-readable data formats (e.g., English sentences or structured code snippets). In LLM4graph, it specifically denotes the textual representation of graph information, encompassing both natural language and structured programming languages. These formats guide the model in analyzing the grammar, semantics, and context associated with graph structures.
    \item \textbf{Tokens} are atomic units of text, (e.g., subwords, characters), often converted to integers via a tokenizer. In LLM4graph, tokens are numerical representations derived from auxiliary neural network processing, including embeddings of graph nodes, node pairs, subgraphs, or entire graphs.
\end{itemize}

Prompts and context together serve as vital enhancements for the input to LLMs. For our two kinds of inputs, prompts can take diverse forms, including questions, descriptions, or task-specific instructions. \textbf{Prompts} guide the LLM’s behavior by specifying instructions or queries. \textbf{Context} refers to the continuous sequence of inputs the model considers when generating a response, influencing the model's capacity to utilize relevant information effectively \cite{zhu2024can}. The length and quality of this context influence how effectively LLMs captures relevant details.

LLMs typically follow three stages to generate output:
\begin{enumerate}
    \item \emph{Tokenization}: Splitting the input text into smaller units called tokens. Each token is then mapped to a unique integer index.
    \item \emph{Embedding}: Transforming the indices into continuous vector representations. This process captures semantic information about the tokens, allowing the model to understand relationships between them.
    \item \emph{Generation}: The model predicts the next token in the sequence step-by-step, using the context provided by the preceding tokens.
\end{enumerate}
When the input is already in a tokenized or embedded format, parts of the initial pipeline may be bypassed or adjusted.
\subsection{Graph Encoders for Large Language Models}
\label{encoder}
Graph encoders transform the structure of a graph into representations (e.g., vectors or sequences) that LLMs or other models can process directly. Common encoder approaches include, attention mechanisms, Sequence Models such as Recurrent Neural Networks (RNN) \cite{sutskever2011generating}, Long Short-Term Memory (LSTM) \cite{graves2012long}, or Gated Recurrent Units (GRU) \cite{chung2014empirical} and GNNs such as GCN \cite{kipf2016semi}, GAT \cite{velickovic2017graph} and GIN \cite{xu2018powerful} are often employed to directly encode graph structures into embeddings. Additionally, LSTM and Transformer-based decoders are also applied after representing graphs as sequences.

\paragraph{Attention Mechanisms.}
In the core prediction phase of the LLM, the attention mechanism calculates the weights of different parts of the input information using Scaled Dot-Product Attention \cite{vaswani2017attention}, enabling the model to assign different importance weights to different segments of the input. Formally:
\begin{equation}
\text{Attention}(\mathbf{Q}, \mathbf{K}, \mathbf{V}) = \sigma\left(\frac{\mathbf{QK}^{\top}}{\sqrt{d_k}}\right)\mathbf{V},
\end{equation}
where
the query matrix $\mathbf{Q} \in \mathbb{R}^{n \times d_k}$ represents the current token's embedding that is seeking contextual information from other tokens,
the key matrix $\mathbf{K} \in \mathbb{R}^{m \times d_k}$ represents the embeddings of all tokens in the sequence,
the value matrix $\mathbf{V} \in \mathbb{R}^{m \times d_v}$ contains the actual information or embeddings of the tokens that will be weighted and summed based on the attention score to produce the output,
and \(\sigma\) typically denotes a softmax function. Here, $m$ and $n$ are the dimensions of the parameter matrices (\textit{n} is the sequence length), $\mathbf{d_k}$ is the dimensionality of the queries and 
keys, and $\mathbf{d_v}$ is the dimensionality of the values.

\paragraph{Graph Neural Networks (GNNs).}
GNNs adopt a message-passing scheme to update node features and learn graph embeddings that preserve the nodes, edges, and structure of the graph. A GNN is formalized as follows: 
\begin{equation}
\operatorname{GNN}(\cdot) = \text{UPDATE}\left(\mathbf{h}_v, \text{AGGREGATE}\left(\left\{\mathbf{h}_u \mid u \in \mathcal{N}(v)\right\}\right)\right) 
\end{equation}
where $\mathcal{N}(v)$ denotes the set of neighboring nodes of node $v$, $\text{AGGREGATE}(\cdot)$ is an aggregation function used to combine neighbor information into a single vector, and $\text{UPDATE}(\cdot)$ is an update function that incorporates neighbor information and node features to iteratively update the node representation.

\paragraph{Sequence Models.}
Once the graph structure is converted into sequence data, the recurrent structure of sequence models can be used to preserve historical information through hidden states.
A multi-layer sequence model can be expressed as:
\begin{equation}
\text{Seq}(\mathbf{x}) = \text{Enc}_N\left(\text{Enc}_{N-1}\left(\ldots \text{Enc}_1(\mathbf{x}_1, \mathbf{x}_2, \ldots, \mathbf{x}_T) \ldots \right)\right),
\end{equation}
where $\text{Enc}_i$ represents the $i$th layer in the $N$-layer encoder (such as an RNN time step or a Transformer encoder block), $T$ is the length of the input sequence, and $x_t$ is the element of the input sequence at time step $t$.

\section{Challenges in Transforming Graphs to Texts}
\label{challenge}

\begin{figure}[h!]
    \centering
    \includegraphics[width=0.8\linewidth]{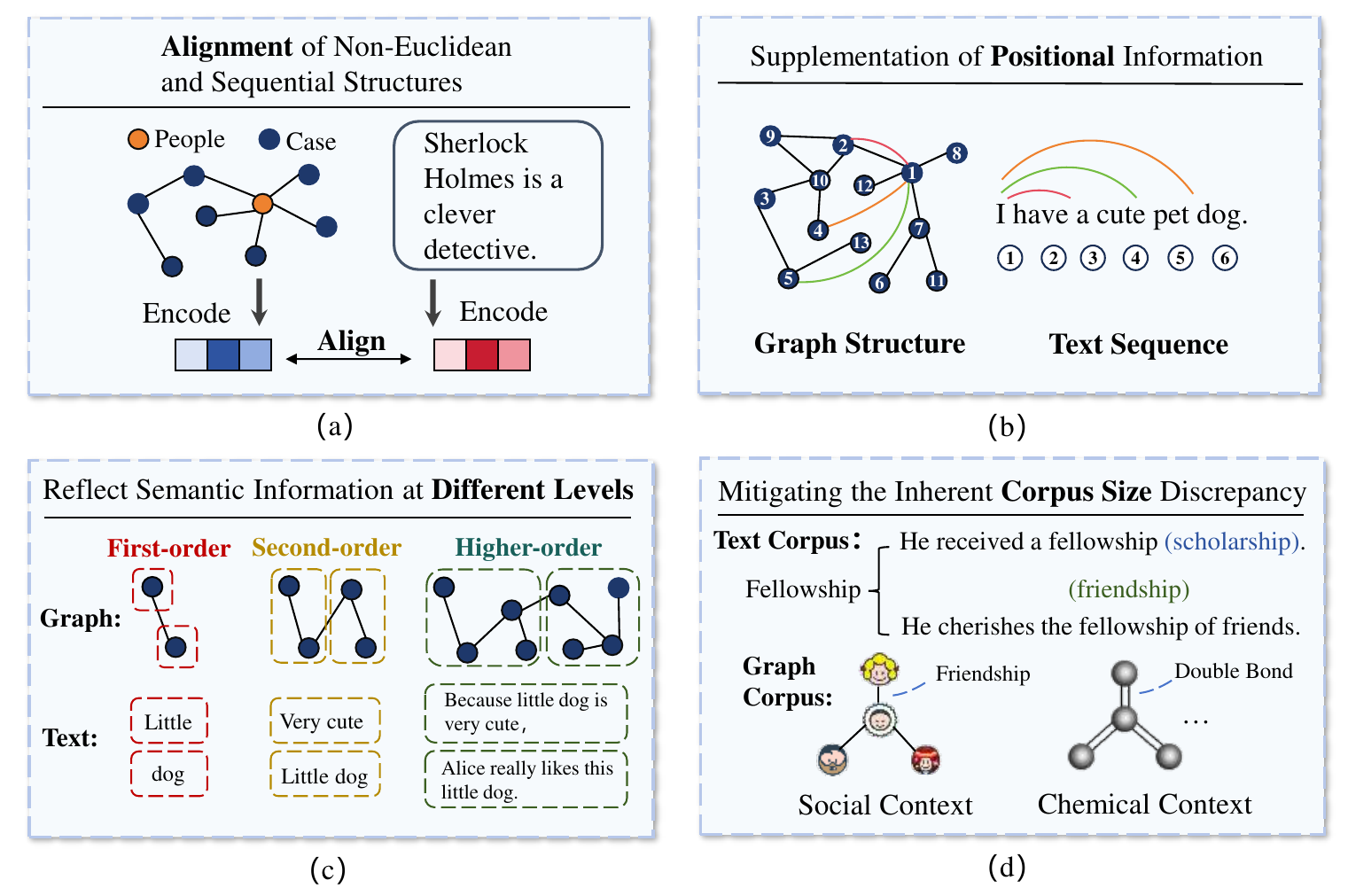}
    \caption{Challenges in transforming graphs to texts. (a) The alignment problem. (b) The position problem. (c) The multi-level semantics problem. (d) The context problem.}
    \label{fig:challenge}
\end{figure}

In this section, we identify four key challenges in transforming graph structures into text.
Since graphs are highly irregular, while texts are regular sequences, texts and graphs have different data structures.
In order to let LLMs comprehend graphs as texts, the transformation must consider and mitigate the crucial differences between these two modalities.
As shown in Figure \ref{fig:challenge}, we focus on four critical challenges: 
\textbf{the alignment problem}, 
\textbf{the position problem}, 
\textbf{the multi-level semantics problem}, 
and \textbf{the context problem}.
We adopt these challenges \textit{as the criteria to subclassify existing methods} in Graph2text (Section \ref{sec:Graph2text}) and Graph2token (Section \ref{sec:Graph2token}).
When the transformation tackles these challenges, the resulting graph representation incorporates the characteristics of texts while preserving the structural information of the original graph.

\noindent\textbf{The alignment problem.}
Traditionally, "alignment" refers to bridge of two modalities \cite{karpathy2015deep}, which is different from transforming one modality (graphs) to another modality (texts) in the scenario of LLM4graph.
However, some graphs (e.g., citation networks) contain abundant textual information that should be aligned with the inherent structural information.
Thus, we adopt "alignment" to generally summarize the process of supplementing/fusing characteristics of graphs and texts.
As shown in Figure \ref{fig:challenge}(a), we state the alignment problem from a macro perspective, which does not consider the characteristics of the two aligned modalities.
For example, when adopting a self-attention mechanism to fuse and align the representation of two modalities, the self-attention does not account for the specific nature of two input modalities, though the characteristics of graphs and texts are preserved. 
Since tackling this alignment problem may not need to consider the specific characteristics of graphs or texts, existing methods (e.g., alignment attentions for texts and images) can be directly used for LLM4graph.
If the alignment problem is overlooked, LLMs may fail to fully capture the interplay between the structural dependencies in graphs and the semantic richness in texts, leading to incomplete or suboptimal representations. For instance, without explicit alignment, the structural connectivity of a citation network might be overshadowed by textual descriptions, making it challenging for LLMs to reason over graph-specific properties (e.g., node centrality).

\noindent\textbf{The position problem.}
The text naturally imposes an order. There are always a "first" and "second" word, while graphs lack a canonical node ordering. Thus, we cannot appoint `the first node' in a general graph.
As shown in Figure \ref{fig:challenge}(b), nodes in a graph do not have fixed positions, while words in a text always have fixed positions.
This fundamental discrepancy makes graphs and texts two distinct data structures.
Tackling this position problem involves analogizing distances on graphs (e.g., relative positions through shortest paths) and generalizing it into textual representation (e.g., position encoding).
While there is an explicit positional order between words in a text sequence, the position information from the graph structure must be integrated into texts to bridge this gap.

\noindent\textbf{The multi-level semantics problem.}
This problem indicates that textual representation always has multi-level semantics, which should also be analogized in graphs.
We give three examples to illustrate this multi-level semantics problem.
Firstly, words in texts always have \textit{dependencies at different hops}. For example, in a text sequence, there are short-range dependencies like "\underline{I} am under the \underline{tree}", mid-range dependencies like "\underline{I} often sit under the big \underline{tree}", and long-range dependencies like "\underline{I} am often found sitting under the shade of a \underline{tree}, one that has been growing for many years." 
Similarly, the hop between nodes also has relations in different hops. However, graph-based methods (e.g., GNNs) for modeling node level usually only consider direct connections, i.e., one-hop relationships. 
Consequently, these methods do not explicitly consider relations at different hops and cannot be analogized to texts.
Secondly, texts have varying \textit{semantic levels} present in the text. For example, in the sentence "Because little dog is very cute, Alice really likes this little dog," the words "little" and "dog" have a first-order connection, "little dog" and "cute" form a second-order connection between the phrase and the word, and "Because little dog is very cute" and "Alice really likes this little dog" establish a higher-order connection between clauses. However, the original construction of graphs (i.e., edges between nodes) is only first-order, and graph-based models only consider these first-order relations and overlook high-order relations (e.g., treat a pair of nodes as an element rather than two separate nodes).
Thirdly, text sequences also contain rich positive or negative semantics, such as "I like dogs" and "I \underline{do not} like dogs", which is essential in natural language. This present or absence of simple words `do not' reflect an opposite meaning despite being a similar sentence. The construction of graphs only considers the presence of edges that serve as positive meaning. It is worth noting that there is also the fact that two nodes are disconnected, serving as negative meaning in graphs. However, existing encoding methods only consider the positive existence of edges, neglecting all negative pairs in graphs. When overlooking this problem when transforming graphs to texts, the representation will lack the negative meaning of texts which is crucial to composing human languages.

\noindent\textbf{The context problem.}
Language models learn word meanings by analyzing large, varied text corpora. Text corpora are commonly large datasets, potentially including millions of words or more across various documents.
By contrast, real-world graphs often have relatively few nodes (hundreds or thousands), providing limited “corpus” for a node to develop a rich context.
t is worth noting that texts are always regular sequences, and easier to infer the semantic meanings of words in such a sequence.
However, graphs are highly irregular with a small size, and graphs are always less dense in terms of content.
Thus, graphs do not contain enough structure or relationships to generalize well or to represent complex phenomena effectively.
Additionally, existing graph-based models concentrate on the neighborhood relationships of nodes and do not consider the role of a node in the whole graph context.
As shown in Figure \ref{fig:challenge}(d), a structured pattern in a social network has a different semantic meaning compared to that in a chemical network.
Note that, graph-based models (e.g., GNNs) only concentrate on the structural patterns, but they cannot capture their different semantic meanings in different graph contexts.
Graph-based models typically focus on local structure and do not always encode the domain-specific \emph{context} that shapes node roles. 
To address this, techniques such as data augmentation, domain prompts, or explicit corpus expansion may be required to convey richer contextual information about each node or substructure.
 

\section{Graph2text}
\label{sec:Graph2text}
In this section, we introduce Graph2text, the process of converting graph data into textual descriptions suitable for LLMs. We categorize Graph2text methods into three main types based on the graph’s application domain: \emph{general graphs}, \emph{AMR graphs}, and \emph{knowledge graphs (KGs)}.

\begin{figure}[h!]
    \centering
    \includegraphics[width=1\linewidth]{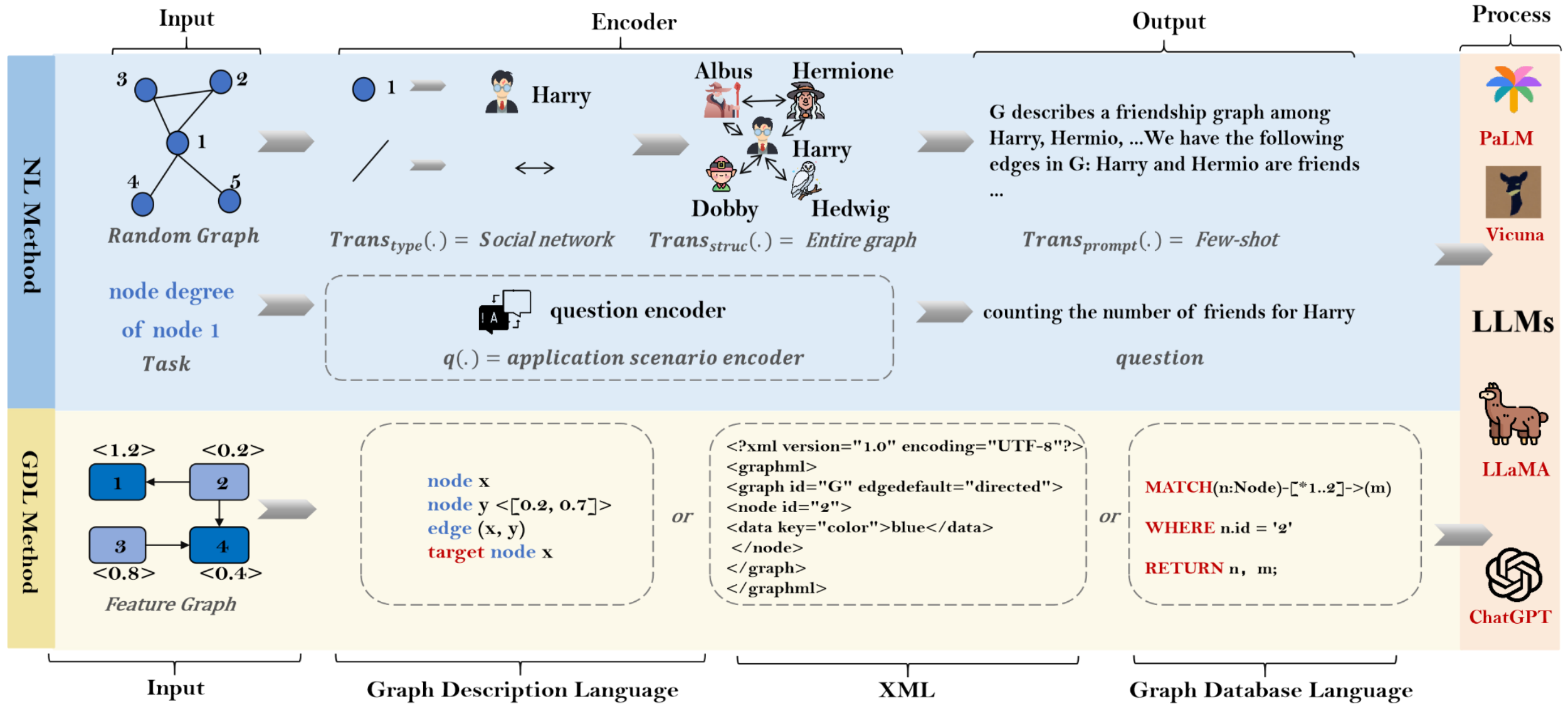}
    \caption{Direct textualisation with Natural language generation and GDL generation.}
    \label{FIG3}
\end{figure}

\begin{figure}[h!]
    \centering
    \includegraphics[width=1\linewidth]{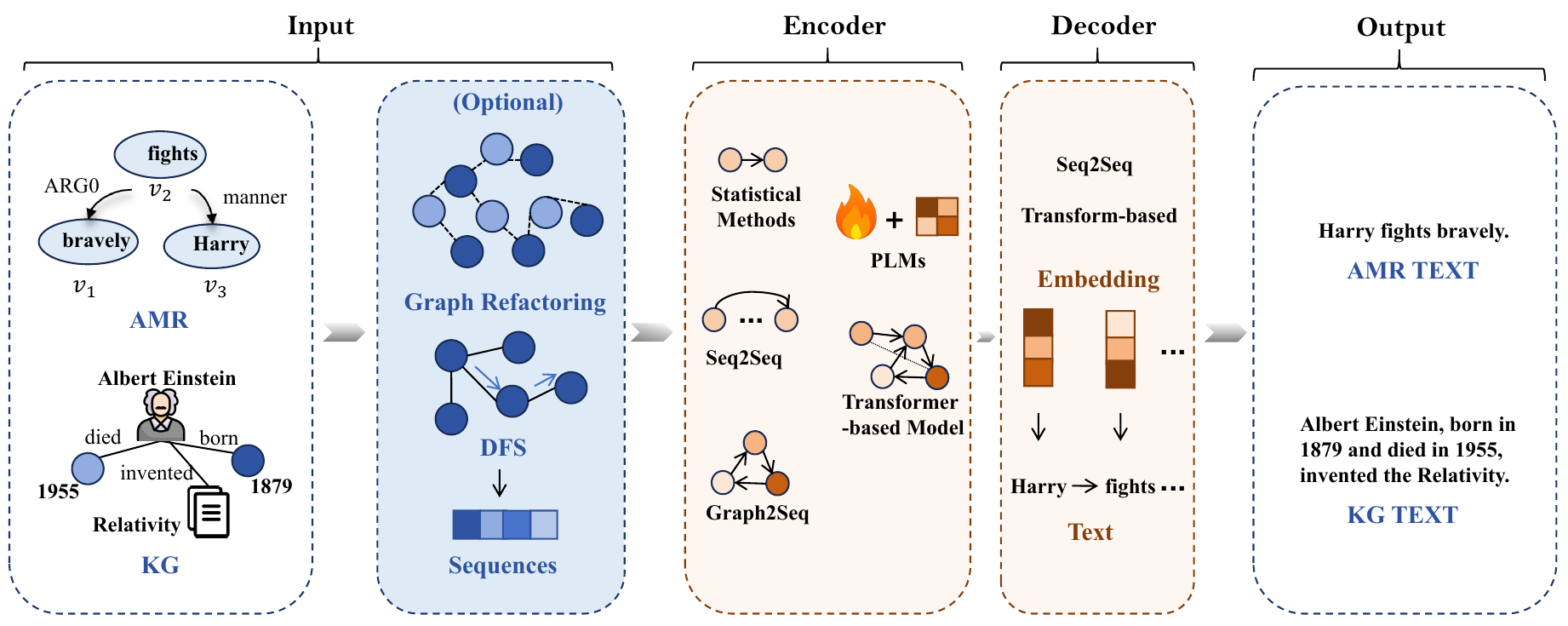}
    \caption{Technical approach of the indirect textualization method.}
    \label{FIG4}
\end{figure}

For \emph{general graphs}, the key challenge is designing a flexible framework that can handle any domain (e.g., biological, social, or citation networks) by choosing a proper granularity (nodes, edges, subgraphs) and appropriate text-generation strategies.  As shown in Figure \ref{FIG3}, the general graph method uses natural language to describe the structure of the graph and adjusts the generated text based on the type of network. This process directly modifies the expressions of nodes and edges or adds domain-specific knowledge as a supplement. Since \emph{AMR graphs} serve as intermediate states that abstractly represent natural language text and \emph{KGs} can directly act as knowledge sources, Graph2text applications are predominantly found in AMR and KG domains. Given that these graphs have predefined structures, it is easier to design specialized encoders for converting these graphs into text. As depicted in Figure \ref{FIG4}, the encoder conversion process involves six steps: Graph Serialization, Encoder Processing, Hidden State Extraction, Decoder Generation, Probability Distribution, and Next Word Prediction.

\raggedbottom

\raggedbottom

\subsection{General Graphs: Transforming Structures into Textual Representation}
The key to the general graph method is choosing the appropriate granularity for describing graph structures based on the graph type. (e.g., nodes, subgraphs, or entire graphs) and selecting suitable conversion functions. For instance, in biological networks, nodes may represent molecules and edges may denote chemical bonds. This approach enables converting graphs into text, which can either fine-tune LLMs or serve as prompts to assist LLMs in handling graphs from various domains, ensuring that the final LLM-generated answer is as consistent as possible with the expected outcome.

General graph methods are categorized into Natural Language (NL) Method and Graph Description Language (GDL) Method based on the format of the transformed text. The NL Method explicitly generates natural language descriptions of graph structures for specific tasks (e.g., node degree, topological sorting), which are often used as prompts for LLMs. The GDL Method employs specialized labels or syntactic rules to organize the text more structurally, aiding LLMs in better understanding the prompts for graph-related tasks. Both methods mainly address the alignment problem of converting graphs into text. 

We define the architecture of the general graph method can be summarized as:
\begin{equation}
S = \text{Trans}_{type}(\text{Trans}_{struc}(G), \text{Trans}_{prompt}(T_{domain})),
\end{equation}
where $S$ represents the text sequence generated by converting the graph using the general graph method. $\text{Trans}_{type}(\cdot)$ refers to the transformation of the appropriate conversion function according to the graph type. $\text{Trans}_{struc}(\cdot)$ refers to the choice of graph granularity for text conversion. $T_{domain}$ refers to the examples or templates of domain knowledge. $\text{Trans}_{prompt}(T_{domain})$ refers to the selection of prompt techniques, such as few-shot learning \cite{brown2020language} or chain-of-thought \cite{wei2022chain}, which incorporate domain-specific knowledge as examples to guide LLMs.

The most straightforward way to convert a graph into text is through natural language descriptions. As shown in Figure \ref{FIG3}, GraphQA \cite{fatemi2023talk} designs a graph embedding module that abstracts descriptions of edges or nodes into encoding functions tailored to specific tasks. Based on actual problems, this module uses semantically related functions to convert edges and nodes into text, which are then combined into a complete prompt.

The performance of NL methods heavily depends on the quality of the prompts. One line of research, similar to GraphQA, focuses on designing effective manual prompts. In GraphQA, the nodes and edges of the graph are converted into text, and then simply combined into prompts. However, this approach neglects the structural information within the graph. To enable LLMs to better understand graph structures before solving related problems, NLGraph \cite{wang2024can} provides algorithmic descriptions in prompts that extract the graph’s structural features. For example, the model incorporates "Build-a-Graph" prompting, which requests actions such as "Next, I want to build a graph with nodes and edges." It also uses algorithmic prompting to guide detailed steps for depth-first search (DFS) traversal. Previous prompt technologies often conflated spatial and temporal dimensions, leading to LLMs' inability to fully leverage time-varying topological structures. LLM4DyG \cite{zhang2023llm4dyg} introduced Disentangled Spatial-Temporal Thoughts (DST2), where time information is prompted first, followed by node position information, helping LLMs better understand dynamic graph structures. However, manual prompting often relies too much on human design and external resources, making it difficult to adapt the generated prompts to the actual problems encountered by the LLM, leading to subjectivity and limited prompt optimization.

Another line of research involves self-prompting and API-prompting, which aim to improve the reusability and scalability of prompts for LLMs. Self-prompting leverages LLMs' learned knowledge to automatically generate new prompts, facilitating knowledge reuse. For instance, as shown in the lower half of Figure \ref{FIG3}, Gpt4graph \cite{guo2023gpt4graph} inputs GDL into LLMs and asks questions related to graph structures, generating new contextual information like structure summaries. This information is then fed back into LLMs, fostering iterative knowledge enhancement. However, automatically generated prompts often lack interpretability and problem-solving feasibility. Moreover, the requirements for the model to generate prompts are also rigorous. They need to meet certain self-learning and optimization capabilities to generate new prompts.
API-prompting allows LLMs interfaces, like ChatGPT, to be called directly, facilitating the learning of graph structure-related examples for prompt generation. This process requires less detailed knowledge of the model's internal workings and imposes fewer demands on both human designers and the model itself. Graph-ToolFormer \cite{zhang2023graph}, an example of API-prompting, calls ChatGPT to generate sentences incorporating external graph reasoning APIs based on graph reasoning examples. However, the reliance on external graph reasoning APIs for training datasets limits the flexibility of LLMs, making them less adaptable to solving specific problems based on the generated prompts.

LLMs are uncertain to understand graph structures in natural language. Thus, various benchmarks have been proposed to verify its ability to solve graph theory tasks. In contrast,  LLMs’ ability to recognize and generate programming languages is nearly undisputed. This is because LLMs are trained on vast amounts of programming data, while the corpus related to graph theory and graph reasoning, especially in natural language, is relatively limited. Moreover, when generating solutions, LLMs often lack responsibility and explainability due to their limited understanding of graph structure. This limitation makes it challenging for LLMs to solve graph theory problems, which require precise and verifiable outcomes. To address this shortfall, the concept of graph-to-programming languages has emerged. Graphs, being highly structured data, present a challenge when described using loose, unstructured natural languages. However, employing programming languages that share strong logical and structural properties can mitigate this discrepancy in Graph2text conversions. Consequently, the GDL Method helps LLMs fully comprehend graph structures before tackling graph theory-related problems.

PL4XGL \cite{jeon2024pl4xgl} uses unified syntax and semantic rules to constrain each GDL program, applying multiple GDL programs to describe the graph features of the same graph component (node, edge, or subgraph) from different perspectives. It then selects the GDL program that best reflects the structural information of the graph component to complete the conversion from graph to text, as illustrated in Fig \ref{FIG3}. However, the custom and fixed format of the GDL program limits its versatility when converting graphs into text.
A GDL program is a planned, linear language description that employs fixed labels, such as "<node>" or "<edge>", to enumerate nodes or edges in the graph. This indiscriminate linear enumeration makes it difficult to capture the true structure of graph data, inevitably leading to the partial loss of structural information. By contrast, the Extensible Markup Language (XML) is organized in a tree structure. Since the graph can also be represented as a tree from any node, so XML helps to represent the structure of the graph. 

In addition, XML is a mature language system that is likely to have been well-represented in the training corpus of LLMs. Therefore, compared to custom GDL, LLMs have stronger capabilities for processing XML. To fully leverage the advantages of XML, as shown in Fig \ref{FIG3}, GraphML \cite{brandes2013graph} uses XML syntax to convert graphs into programming languages. This method does not rely on a fixed set of labels, instead, it uses flexible label-setting rules, allowing it to handle different types of graphs—with or without attributes, text, edge types, and more—when converting them to text.

However, as a markup language, XML displays all information, including the graph’s structure, node types, edge types, and more, in a tree-like format. While this fully presents the structural information, when a graph becomes too large, the converted XML code not only becomes lengthy due to auxiliary elements like labels but also grows increasingly complex in structure. This lengthy and intricate format can obscure important details of the graph structure, making it more difficult for LLMs to understand the graph. To address this issue, our proposal is to suggest using graph database languages such as Cypher \cite{francis2018cypher}, RAT-SQL \cite{wang2019rat}, and Gremlin \cite{rodriguez2015gremlin}. The suggestion is to store the graph in a graph database first, and then use different graph database query statements to retrieve the graph structure information in the graph database and output it. The output will be merged with the input accepted by LLMs to guide LLMs to grasp the graph structure more effectively. The benefits of this method are as follows:

First, declarative graph database query statements concisely and efficiently guide LLMs on how to understand the structure of the graph. As illustrated in Fig \ref{FIG3}, in the query statement input, "MATCH" explicitly tells LLMs to find the conditions that the structure must meet (vertices, edges, or subgraphs) within the graph, while "WHERE" defines the filtering conditions based on graph attributes. The "RETURN" clause specifies the format of the structure generated by the query. Providing LLMs with both the "input + output" from these queries enables a more thorough understanding of the graph's structure.

Second, different graph database query statements can be used to describe the structure surrounding the same node, such as the subgraph structure. Graph database query statements allow LLMs to categorize and interpret different types of graph information, reducing ambiguity and improving the clarity of the model’s understanding.

Third, for more intricate graph structures, such as multigraphs or dynamic graphs, these graph database query statements can also deal with them. Concretely, all graphs can be stored in different tables in the graph database separately, and then multi-table join queries can be used to solve complex graph structures by processing logical associations within SQL queries.

Fourth, graph database query statements are specifically designed for managing graph data, providing a robust and standardized way to describe various types of graphs (e.g., directed or undirected). These languages are not only well-suited for handling diverse graph types but can also serve as training corpora for LLMs. Additionally, Gpt4graph \cite{guo2023gpt4graph} has demonstrated that GDL can be understood by LLMs and applied to specific tasks.

\subsection{AMR Graphs: Transforming Abstract Semantics into Natural Language}
The key to the AMR graph method is learning the position information between nodes and edges within the AMR graph, aligning it with the semantic information between concepts and relationships in the sentence to generate fluent text.
\subsubsection{Alignment}
\ 
\newline
To address the issue of Graph2text alignment, JMAR \cite{flanigan2016generation} first simplifies the AMR graph into a spanning tree with a simpler structure and less ambiguity. A series of synthetic rules were then designed to constrain the specific details involved in converting the graph into text. Pourdamghani et al. \cite{pourdamghani2016generating} employed a greedy algorithm to optimize the AMR graph. This algorithm recursively removes semantically irrelevant edges, bringing semantically related concepts closer in the generated text, and ensuring semantic coherence. Manning et al. \cite{manning2019partially} designed specific rules for initial alignment, but challenges arose as a single node could represent different parts of speech (e.g., nouns, adverbs) or grammatical tenses (e.g., progressive tense) in a sentence. To eliminate ambiguity and improve alignment, their approach generates multiple string forms for the same concept and uses a language model integrated with external knowledge to select the best representation.

To use specialized encoders to convert graphs to text, we define the architecture of AMR graph method and KG method can be summarized as:
\begin{equation}
\begin{aligned}
&S = \text{Seq}(\text{Attention}(\text{Enc}(x_i+x_{pos}))) \\
&p(\mathbf{y}_t \mid h_t) = \sigma(\mathbf{W}_{vocab} h_t).
\end{aligned}
\end{equation}
Here, $x_i$ represents the node embedding of the $i$-th node, while $x_{pos}$ refers to the positional embeddings. $S$ represents the text generated by the graph transformation. $\text{Enc}(\cdot)$ is the encoder of the neural model, which can be a sequential model or graph-based model. $\text{Seq}(\cdot)$ is the decoder responsible for generating the text sequence. The probability distribution $p$ reflects the likelihood of selecting each word in the vocabulary within the current context. The matrix $\mathbf{W}_{\text{vocab}}$ acts as a bridge between the hidden representation of the model $h_{t}$ and the vocabulary, enabling the model to predict the next word by assigning probabilities to all possible tokens in the vocabulary. The subscript $vocab$ emphasizes this mapping to the vocabulary space, and $\sigma$ is the softmax activation function.

For the alignment problem, Neural AMR \cite{konstas2017neural} first uses the DFS algorithm and then declares the type or adjacent structural information after the node and edge to make full use of the AMR's own graph structure information. However, the original AMR focuses solely on deep semantic information, without accounting for grammatical structure or vocabulary forms. Consequently, the AMR graph uses a simple acyclic directed graph to represent semantic flow, increasing structural rigidity. For example, "Alice likes Bob" can only create a directed relationship from "Alice" to "Bob" without accounting for the reverse. This strong structural orientation ignores surface syntax and semantics, making text alignment more difficult. To address this, Beck et al. \cite{beck2018graph} introduced reverse edges and self-loops to the AMR graph, aiming to add surface syntax and semantic potential to the structure, thereby allowing for richer word interactions. Additionally, the model uses a Levi graph to transform edges into unified node expressions, weakening the rigid structural assumptions of the AMR graph. The GNN-RNN encoder-decoder architecture is then used to achieve final alignment. Mix-Order GAT \cite{zhao2020line} further divides the AMR graph into a concept graph and a relation graph, training them separately in the encoder. Some concept graph nodes are not directly connected but have indirect semantic connections, which the relation graph supplements. Therefore, a cross-graph communication mechanism based on attention is proposed. This mechanism selectively focuses on key information from both the concept and relation graphs, facilitating better alignment between the text's semantic meaning and the graph's structure.

Due to the pre-training of large datasets, Pre-trained Language Models (PLMs) possess significant capabilities for encoding graph structures and generating text sequences. To further enhance their alignment performance, SPRING \cite{bevilacqua2021one} optimizes the linearization process, which refers to the process of flattening a graph into a sequence, to ensure that the converted sequence more effectively preserves the integrity of the graph structure. Bevilacqua et al. \cite{bevilacqua2021one} introduced a set of special tokens to mark the same node in the AMR graph when it is pointed to by multiple edges, enabling different textual representations for the same entity. Additionally, since the original vocabulary in BART \cite{lewis2019bart} is optimized for general English text, it modifies the vocabulary to generate text that is more semantically aligned with AMR graphs.
\subsubsection{Position}
\ 
\newline
To address the positional problem, JMAR \cite{flanigan2016generation} uses a transducer input representation to perform lexicographic sorting based on the attributes of the AMR graph. This sorting is also applied when converting the graph to a spanning tree. Although the simplicity, this approach provides a fixed order for graphs that lack a predefined node order. However, relying on traversal combined with lexicographic sorting oversimplifies the process, leading to the loss of structural information that could better organize the nodes.

To mitigate this, Pourdamghani et al. \cite{pourdamghani2016generating} proposed first converting the AMR graph into a tree and assigning a value to each node or edge. This value represents the median of the position values of all nodes or edges within the connected subtree, offering an approximate position for the edge in the generated English sentence. Before text generation, the model sorts the values of all sibling edges in the AMR graph, thus obtaining structural information between neighboring nodes.
Manning et al. \cite{manning2019partially} used graph models for structural and relational analysis, including a pair-order model and a coreness model, to determine the positional relationship between parent and child nodes within a sequence. The placement is based on the text concept corresponding to the node or edge in the AMR graph. For instance, in the coreness model, an "ARG1" node, typically representing an object, is placed closer to the parent node than a "time" node, which often functions as an adverbial.

Neural AMR \cite{konstas2017neural} addresses the node positioning issue not by adjusting node order but by introducing position sharing for similar nodes. For example, entities with similar semantics, such as "country" and "city," are abstracted into the same type of nodes in the AMR graph. This method leverages the knowledge in the text to provide position information for nodes in the graph. Beck et al. \cite{beck2018graph} further improved position handling by incorporating positional information into node embeddings before employing alignment.

While neighboring nodes in the graph only offer local positional information, the learned structural data remains limited. To enhance this, Graph Transformer \cite{cai2020graph} designs a sequence encoder to encode the shortest distance between nodes as relative positional embeddings. The relative position embedding $r_{ij}$ and relative position $x_{pos}^{i \rightarrow j}$ are defined as:
\begin{equation}
\begin{aligned}
&r_{ij}=\text{Seq}(h_{i}, x_{pos}^{i \rightarrow j}) \\
&x_{pos}^{i \rightarrow j}=\left[e(i, k_1), e(k_1, k_2), \cdots, e(k_n, j)\right].
\end{aligned}
\end{equation}
Here, $Seq(\cdot)$ refers to using $GRU(\cdot)$ to calculate the final hidden state $h_{i}$ from both forward and backward directions. $e(\cdot, \cdot)$ represents an edge, and $k_1, \ldots, k_n$ are the intermediate nodes along the shortest path. 

Additionally, the minimum distance from each node to the root in the AMR graph is added as an absolute positional embedding to the nodes’ encoded embedding. Graph Transformer then uses global attention to focus simultaneously on absolute and relative positional data, generating node representations that are passed to the decoder to produce text. However, the vocabulary used to train the model is only a limited subset of the target language. To mitigate vocabulary sparsity during the decoding stage, Graph Transformer employs a copy mechanism, allowing the model to copy low-frequency words directly from the input AMR graph’s concept nodes. The probability distribution is updated as follows:
\begin{equation}
p(y \mid h_t)=p_{copy} * \alpha_{copy}+p_{vocab} * \alpha_{vocab}.
\end{equation}
Here, $p_{\text{copy}}$ represents the likelihood of the model copying a word with the same surface semantics, while $p_{\text{vocab}}$ represents the likelihood of generating a word. The final prediction probability $p(y\mid h_t)$ combines the multi-head attention mechanism with the copy mechanism, resulting in a weighted sum of the generation and copy probabilities. $\alpha_{copy}$ and $\alpha_{vocab}$ are the weights corresponding to the two probabilities respectively.
\subsubsection{Hierarchy}
\ 
\newline
In graph-based structures, the natural ordering exists only between words, forming a first-order structure. However, focusing solely on first-order structures can overlook the varying semantic levels presented in the text. For example, in the sentence "Because dogs are cute, Alice likes the dog very much," the words "very" and "much" have a first-order connection, "likes" and "very much" form a second-order connection between phrases, and "Because dogs are cute" and "Alice likes the dog very much" establish a higher-order connection between sentences. Capturing these higher-order relationships is essential for bridging the semantic layers in Graph2text conversion. 
To mimic these high-order semantics in natural language, Graph2text models should go beyond pair-wise relation (i.e., edges) between nodes to higher-order structures \cite{morris2019weisfeiler} on graphs.
However, GNNs and sequence models cannot capture this because they can only capture the first order.
To address this, Mix-Order GAT \cite{zhao2020line} considers different orders of graphs, representing varying levels of semantic complexity, and then uses GAT to adaptively mix information from these different orders. The encoder $\text{MixGAT}$ is defined as follows: 
\begin{equation}
\text{MixGAT}(x_i, R^K)=\text{Concat}\left[R^1(x_i), R^2(x_i), \cdots, R^K(x_i)\right].
\end{equation}
Here, $R^k$ represents the set of neighbor nodes from $1$ to $K$ order. $R^k(x_i)$ represents the set of all node representations that can reach $x_i$ within $k$ hops. The $\text{Concat}(\cdot)$ operation represents the concatenation of the hidden states of all other nodes between $1$ hop and $K$ hop. 
\subsubsection{Context}
\ 
\newline
Guo et al. \cite{guo2019densely} first converted the AMR graph into a Levi graph \cite{beck2018graph}, treating both nodes and edges as components for learning representation. Although the Levi graph can treat edges as nodes and use an encoder to learn the hidden information corresponding to the edge labels, enriching the generated text content, reconstructing the connections in the graph may cause the new graph to not be completely isomorphic to the original. This discrepancy can affect the information exchange between nodes that were directly connected in the original graph, leading to a loss of contextual information in the generated text. Building on this Levi graph, a virtual node \cite{gilmer2017neural} is introduced that connects all nodes. This global node acts as a transfer station for all node representation information, consolidating the graph's context and distributing it to all connected nodes. This compensates summarize context of the whole graph for every node, thus addressing the lack of contextual information in Graph2text tasks.

\subsection{Knowledge Graphs: Transforming Structured Knowledge into Natural Language}
The key to KG method is determining the connection position between entities and relations, which includes the order of entities within facts, their expression in sentences, and the sequence of sentences themselves \cite{moryossef2019step}. This process aims to align Graph2text properly to ensure that the structured knowledge is being inaccurately converted, avoiding missing information or even logically flawed content.
\subsubsection{Alignment}
\ 
\newline
WalkLM \cite{tan2024walklm} utilizes a random walk to sample nodes and edges from the graph. The text corresponding to the sampled sequence is then used to fine-tune the pre-trained language model. This approach aligns graph structure information with text semantics, transferring knowledge from the pre-trained model to a specific domain through knowledge distillation, thereby providing structural knowledge to LLMs.

The study by Marcheggiani and Perez-Beltrachini \cite{marcheggiani-perez-beltrachini-2018-deep} employs a graph-based model (i.e., GCN) as the encoder and a sequence-based model (i.e., LSTM) as the decoder. Here, the structural information is first encoded into the sequence, which is then decoded to obtain alignment. During the linearization process, the model incorporates structural information obtained from DFS traversal into the sequence of tokens derived from the KG. To achieve proper alignment, GraphWriter \cite{koncel2019text} unifies the nodes and edges to the same level. It converts edges into two nodes to represent both the forward and reverse directions of the original relationship (edge), effectively reconstructing the graph into a new bipartite structure. This method relaxes the strict assumption of structures about the KG graph while preserving the original structure, facilitating alignment between the graph and text modalities. Additionally, the KG content’s topics are processed through a Title Encoder, ensuring that both text knowledge and graph structure encoding are combined via an attention layer for enhanced alignment.

GAP \cite{colas2022gap} encodes the mutual attention relationships between all entities and relations in the graph via a generalized adjacency matrix, capturing domain-specific structural information. In this way, the structural information encoded by the graph encoder is considered, and the LM is pre-trained with the linearization sequence, allowing each node or edge to retain both structural and textual information, thus achieving further alignment. JointGT \cite{ke2021jointgt} introduces a component called the Structure-Aware Semantic Aggregation Module, which uses self-attention to focus on shared position information between each entity/relation in the linearized KG sequence and the auxiliary text describing the KG structure. Furthermore, Ke et al. \cite{ke2021jointgt} designed three enhanced tasks for graph-text alignment: generating the most relevant words for masked text, generating the most appropriate entities or relations for masked KGs, and minimizing the distance between text and graph embeddings to fine-tune the model and ensure alignment completeness.

One line of research has focused on increasing the diversity of pre-training data for alignment. They apply unsupervised training methods to perform both Graph2text and Text2graph conversion tasks using non-parallel graph-text data to achieve bidirectional alignment. Jin et al. \cite{jin2020genwiki} introduced GenWiki, a dataset of 1.3 million text and graph examples, establishing a new benchmark for future research. Similarly, CycleGT \cite{guo2020cyclegt} uses fully non-parallel graph and text data as a training set, addressing data scarcity issues in iterative training tasks. Schmitt et al. \cite{schmitt2020unsupervised} proposed two rule-based systems and an unsupervised training method that automatically adapts to various domains and graph schemas.

\subsubsection{Position}
\ 
\newline
WalkLM \cite{tan2024walklm} employs random walks on the graph when converting the graph into a sequence. The sampled sequence captures the connection position information of each entity and relation. The model then converts the entities or relationships within the sequence into text, concatenating them in a fixed procedure. It is important to note that the random walks are not arbitrary. They are strategically employed to ensure that the generated sequence is rich in positional information. Each step in the walk is a deliberate move that captures the connection points, where entities and relations intersect within the graph. This approach maps out the nodes and edges not only by their existence but by their specific roles and locations within the overall graph architecture, making WalkLM a powerful tool for producing more accurate and nuanced graph embeddings for a variety of graph-related tasks.

\subsubsection{Hierarchy}
\ 
\newline
In a KG, the relationships between entities are diverse, with different types of edges representing varying connections. Since the connection type between the subject and object within a triple is explicitly stated by the predicate, the structural information of intra-triple relationships is relatively easier to encode. To address the information gaps in inter-triple relationships, the GTR-LSTM \cite{distiawan2018gtr} first reconstructs the graph into a strongly connected graph using topological sorting and DFS. This approach simplifies complex KG relationships, such as loops and intra-triple relationships, making it easier to capture multi-level semantic information corresponding to various relationship types with customized LSTM units. The model can then aggregate information from both entities and their directly adjacent edges, allowing the model to learn the specific semantics of all relationships between entities and entity pairs.

In Marcheggiani and Perez-Beltrachini's work \cite{marcheggiani-perez-beltrachini-2018-deep}, the model applies different weight matrices for different types of edges (e.g., varying relationship types and directions) when aggregating information using a GCN. This differentiation of edge types enables the model to capture multiple semantic relationships within the context.
\subsubsection{Context}
\ 
\newline
To further capture the meaning of each entity or relationship across different contexts, Ribeiro et al. \cite{ribeiro2020modeling} leveraged GAT to aggregate information from "neighbor" nodes in the graph, using this information as the semantic expression of the central node in its local context. The model captures context at both global and local levels. Specifically, they achieve this by using local node encoding to compute local contextual information and global node encoding to obtain the global semantic representation of an entity.

As mentioned earlier, GraphWriter \cite{koncel2019text} reconstructs the graph into a bipartite graph for alignment. However, this reconstruction disconnects the connections in the original KG, which hinders communication of context between nodes. To resolve this, the model introduces a global context node in the bipartite graph, enabling each node to access global information, thus addressing the context problem.

To capture multiple semantic relationships, such as subject-predicate or verb-object structures, GAP \cite{colas2022gap} adjusts the attention bias for different entity-relationship connections through Connection Type Encoding. This mechanism effectively captures the subject-predicate-object structure of sentences, addressing the challenge of encoding multiple semantic relationships. The core formula for adjusting attention to capture different semantics within a context is as follows:
\begin{equation}
\operatorname{Attention}_{M, T}(\textbf{Q}, \textbf{K}, \textbf{V})=\operatorname{softmax}\left(\frac{\textbf{Q} \textbf{K}^{\top}}{\sqrt{d_{k}}}+\textbf{M}+\textbf{T}\right) \textbf{V},
\end{equation}
where the mask matrix $\textbf{M}$ is a generalized adjacency matrix used to indicate whether entities or relationships should pay attention to one another. The type encoding matrix $\textbf{T}$ assigns constant values ranging from 0 to 4 for different types of connections between entities and relationships (e.g., entity-entity, relationship-relationship). Additionally, the model linearizes the graph into word tokens for pre-training a language model, allowing each token to predict its role in the global context and initially learn the contextual information.

\section{Graph2token}
\label{sec:Graph2token}
This section delves into the existing Graph2token methods and technologies. Tokens serve as the fundamental input units for LLMs, and their types and quality significantly influence LLMs performance \cite{chen2024exploring}. The application of LLMs to graph-based tasks presents inherent challenges, as these models are not designed to process graph structural data directly.

\begin{figure}[h!]
    \centering
    \includegraphics[width=0.95\linewidth]{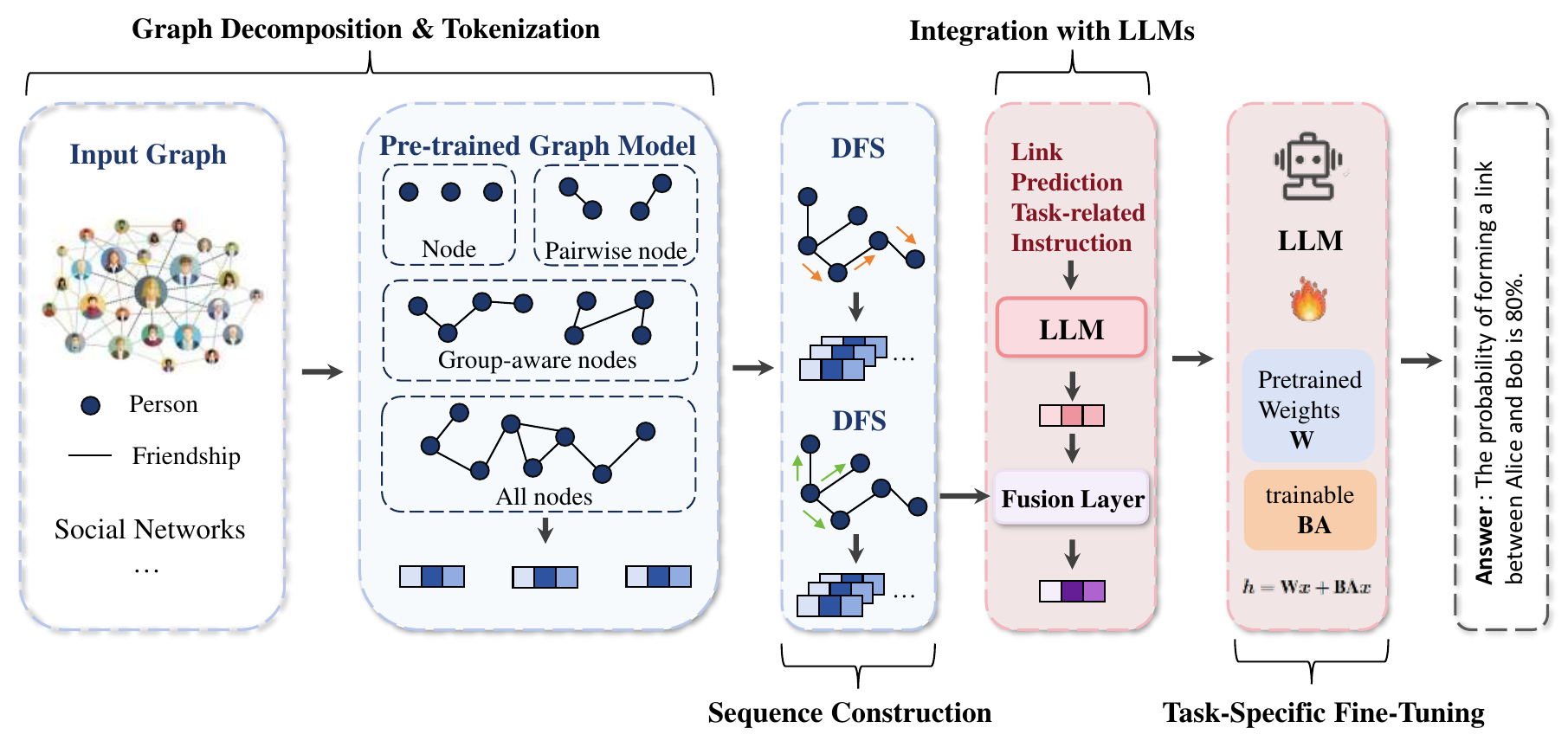}
    \caption{Schematic diagram of Graph2token. Converting graph data to tokens in the graph task to help the large language model understand the processing, using social networks as an example.}
    \label{fig:graph2token}
\end{figure}

Early language models segmented and encoded input text into tokens, such as words and subwords. Inspired by this, the fundamental concept of the Graph2token method is to transform nodes, edges, and their attributes or subgraphs within a graph into a series of tokens. These tokens are then fed into LLMs for processing \cite{min2023recent,li2021terapipe}. The process consists of the following key steps, which are shown schematically in Figure \ref{fig:graph2token}.

1) \textbf{Graph Decomposition:} Graph decomposition involves breaking down the graph into fundamental components such as nodes, edges, and subgraphs. This step includes extracting node identifiers, connection relationships, and related weights or labels. By decomposing the graph into smaller components, it becomes easier to tokenize and process subsequently \cite{sun2023large}.

2) \textbf{Tokenization:} Tokenization is the pivotal step in the entire graph2token process, the subsequent classification steps will build upon this fundamental process. Tokenization involves converting each component of the graph into tokens or sequences of tokens. For instance, nodes can be represented by unique identifiers, and edges can be encoded based on their connections and attributes. Encoding methods include adjacency matrix/list encoding, and graph embedding techniques, which generate continuous vector representations of nodes and edges, capturing the structural features of the graph. The goal of tokenization is to retain the semantic and structural information of the graph, allowing LLMs to understand and process this information \cite{wei2024llmrec}.

3) \textbf{Sequence Construction:} Sequence construction involves arranging tokens in a specific order to encapsulate the graph’s information. This ordering is often achieved through traversal methods like DFS or breadth-first search (BFS), which maintain the graph's structural integrity. The challenge lies in organizing the tokenized graph components into a coherent sequence that LLMs can effectively process. DFS highlights hierarchical relationships, while BFS captures broader structural information. By consistently applying these methods, the sequence construction respects the relative positions of nodes in the original graph. Redundancy is minimized by encoding only the necessary structural features of nodes and edges, focusing on the most informative parts of the graph. A well-constructed token sequence allows LLMs to perform more accurate analysis, ultimately improving the performance of graph-related tasks \cite{perozzi2024let,lu2024grace}. 

4) \textbf{Integration with LLMs:} The token sequences are input into LLMs. Pre-trained LLMs are fine-tuned on these sequences to capture the relationships between nodes and edges in the graph. The encoded tokens are aligned with text prompt embeddings in the projection layer before being input into LLMs. Through this integration, LLMs can leverage their powerful language modeling capabilities to understand and analyze graph data. The fine-tuning process ensures the model accurately captures the structural information in the graph data \cite{wang2024human}.

5) \textbf{Task-Specific Fine-Tuning:} According to the required graph-related tasks such as node classification, link prediction, or graph generation, additional training is performed on the model to optimize its performance. Transfer learning techniques can be applied to further improve the model's performance on specific tasks, e.g., few-shot node classification \cite{wang2022graph}, link prediction \cite{sun2022gppt}. Ensuring LLMs achieves optimal performance when handling specific graph tasks enhances its effectiveness and reliability in practical applications \cite{lin2024data}.

\begin{figure}[h!]
    \centering
    \includegraphics[width=0.95\linewidth]{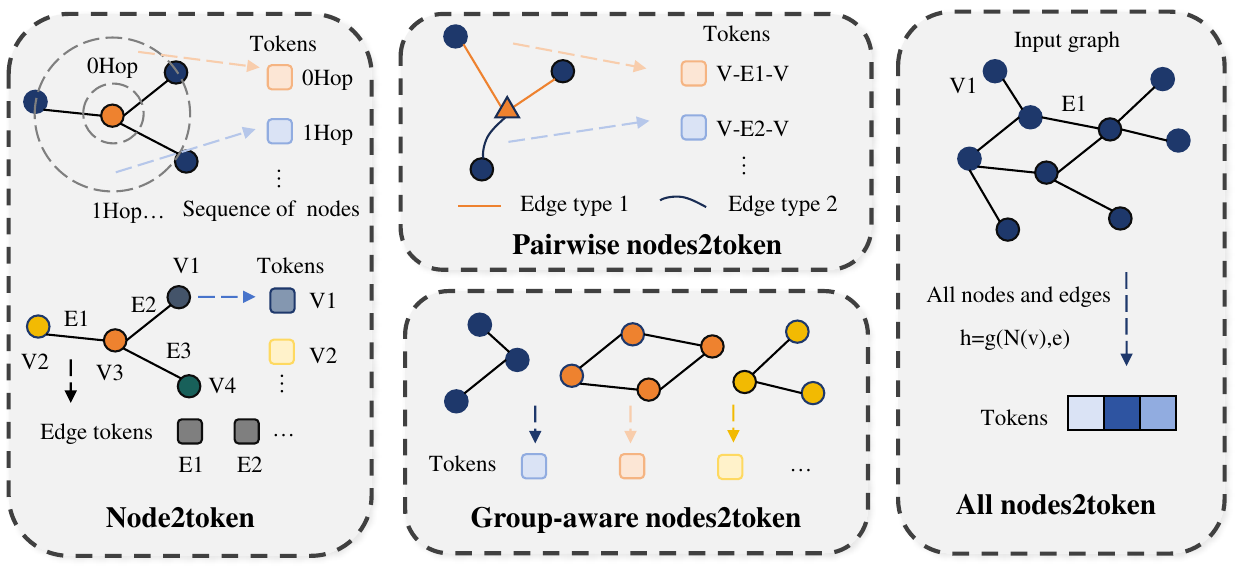}
    \caption{Classification of four tokenization methods based on token granularity.}
    \label{fig:tokenization_methods}
\end{figure}

Tokenization is the cornerstone of the Graph2token method, converting graph data into sequences that LLMs can process directly. This step is pivotal because the type and structure of tokens determine how well the graph's intrinsic information is captured and represented.  \textbf{Given its importance, we have categorized the Graph2token methods into four types based on token granularity during tokenization:  Node2token, Pairwise Nodes2token, Group-aware Nodes2token, and Holistic Nodes2token}, as shown in Figure \ref{fig:tokenization_methods}. The following sections will detail the technical routes and advantages of these methods.

\subsection{Node2token}
Node2token refers to the process of converting nodes that aggregate neighborhood information within graph data into serialized tokens. This method models fine-grained structural information at the node level, distinguishing semantic differences between different nodes by encoding a unique token for each node or edge. These tokens, along with text prompts, are input into LLMs to perform downstream tasks such as node classification and link prediction \cite{wang2024llm,xu2024generate}. This approach is particularly suited for graphs with rich node features and attributes, such as social networks, biological networks, and knowledge graphs. 

\subsubsection{Alignment}
\ 
\newline
Node-level tokenization retains the unique structural representation of each node, integrating the expanded structural information of each node into the corresponding token. These tokens, which form a text-like sequential token series, can be used with textual prompts after the granularity-level structural information is aligned. This process addresses the problem of modality mismatch between graph data and text data \cite{RenMatching2021}. The implementation of this process can be achieved through pre-trained graph models, as shown below:
\begin{equation}
T_i = \operatorname{GM}(h_i) + \text{Align}_{\text{LLM}}(T_i^{\text{text}}),
\end{equation}
where \( T_i \) represents the token of node \( i \), \( \operatorname{GM} \) refers to the well-trained graph model, \( h_i \) is the feature representation of node \( i \), and the \( \text{Align}_{\text{LLM}} \) function is used to align the node's structural token with the language model's pre-existing text token space.

To facilitate the comprehension and processing of graph data by LLMs, it is imperative to align the graph data with textual embeddings prior to ingestion into the model. One effective approach involves representing each node's local neighborhood structure using distinct code indexes, forming a node Token space. This space is then indexed and queried based on the desired information. VQGraph \cite{yang2023vqgraph}, designed with a node dictionary, unifies the knowledge of GNNs and other models. Consequently, VQGraph can collaborate with textual models to efficiently align structural and textual representations.

An alternative solution relaxes the rigid structural assumptions imposed on graphs. TokenGT \cite{kim2022pure}, for instance, treats both nodes and edges as identical tokens, disregarding their functional distinctions. This reduction in structural assumptions mitigates the inductive bias of graphs, enhancing the model's learning capacity to learn graph structures as textual information. As a result, TokenGT not only aligns graph structures with sequences but also adaptively applies diverse graph data to sequence models.

When transforming nodes with rich textual attributes into tokens, aligning the nodes' textual properties is crucial in addition to focusing on their topological structure. The GraphGPT \cite{tang2024graphgpt} framework integrates LLMs with graph structural knowledge through a two-phase command tuning process. It introduces a text-to-graph alignment technique, employing separate graphs and text encoders to encode respective information. By contrasting these encodings, the framework optimizes alignment, effectively aligning graph structures and textual information.

GraphTranslator \cite{zhang2024graphtranslator} proposes a core Translator module that converts structured graph information (e.g., relationships and interests of users A, B, and C) into natural language instructions comprehensible to LLMs. Self-attention and cross-attention mechanisms align the representations of the graph model and language model, facilitating efficient conversion of graphs to textual representation. The Producer module in GraphTranslator further addresses the challenge of aligning graphs without textual descriptions by guiding LLMs to construct descriptive instructions from node, neighborhood, and commonality information.

GraphLLM \cite{chai2023graphllm} proposes an end-to-end method to achieve alignment. It extracts node semantic information using a text Transformer encoder-decoder, aggregates node representations with GTs to generate graph tokens and encodes graph structural information into a fixed-length tunable prefix vector. This prefix vector is integrated with pre-trained LLMs, avoiding inherent information representation gaps due to modality differences.
\subsubsection{Position}
\ 
\newline
The node-level tokenization method introduces explicit node token encoding. This encoding directly incorporates the neighboring information of the node, such as which other nodes it connects with. It can capture and complement the relative positional information of nodes in the graph data, ensuring that position relationships are preserved in textual representations. The encoding process can be expressed as:
\begin{equation}
T_i^{\text{pos}} = T_i + \sum_{j \in \mathcal{N}(i)} \text{PosEncode}(d(i,j)),
\end{equation}
where \( \mathcal{N}(i) \) represents the set of neighbors of the node \( i \), \( d(i,j) \) is the shortest path distance between node \( i \) and node \( j \), and \( \text{PosEncode}(d(i,j)) \) denotes the function that encodes the relative positional information.

For graph data, nodes lack a fixed order. The "order" of different nodes actually corresponds to their relative positions in the graph structure, while text sequences have a fixed order, with both relative and absolute positions. However, the positional information in both graph and text representations is narrow, necessitating a unified positional encoding to simultaneously express positions on both graphs and text. Therefore, in the transformation from graphs to text representations, to enable sequence models to understand and process graph data, a token sequence capable of preserving relative positional information is required to bridge the gap in positional information between graphs and text embeddings. GIMLET \cite{zhao2023gimlet} proposes a transformer mechanism with generalized positional embeddings, employing distance-aware encoding to transform nodes in molecular graphs into tokens. The shortest path distance between nodes is used to calculate positional biases, which are used to generalize the narrow positional information on graphs and text. GraphLLM also uses a similar method to complete position information.

\subsubsection{Hierarchy}
\ 
\newline
The node-level tokenization method assigns specific tokens to each node, preserving the semantic differences of each node. It incorporates a multi-layer aggregation mechanism to perform multi-level analysis, identifying and aggregating information at different levels within the graph. This allows for a deeper analysis of the meaning of each node at different hop levels. For example, some nodes may play a significant role in local structures, while from a global perspective, they may exhibit different relationships and functionalities. The multi-layer aggregation mechanism can be described as:
\begin{equation}
T_i^{(l)} = \text{concat}\left[\{T_j^{(l-1)} \,|\, j \in \mathcal{N}_1(i)\}, \{T_j^{(l-1)} \,|\, j \in \mathcal{N}_2(i)\}, \dots, \{T_j^{(l-1)} \,|\, j \in \mathcal{N}_m(i)\}\right],
\end{equation}
where \( \mathcal{N}_1(i), \mathcal{N}_2(i), \dots, \mathcal{N}_m(i) \) represent different sets of neighbors at increasing hop distances from node \(i\), ensuring that the aggregation captures multi-level neighborhood information effectively.

To capture these simultaneously existing semantic relationships of varying distances in the transformation from graph to text representation, LLaGA \cite{chen2024llaga} introduces node-level templates (neighborhood detail templates and hop-field overview templates). The neighborhood detail template captures the local details of the target node, such as the attributes and connections of directly adjacent nodes, while the hop-field overview template extends to long-range dependencies, enhancing contextual information and effectively capturing dependencies across different distances.

\subsubsection{Context}
\ 
\newline
The node tokenization method reflects the essential properties and functionalities of nodes. It structures them independently of specific graph configurations, allowing these tokens to maintain consistent semantics across different contexts. This method improves the adaptability of nodes to diverse environments. Specifically, introducing an upper and lower context encoding mechanism, which combines the tokens of a node with its surrounding information, creates a rich contextual representation. This process not only considers the node’s own properties but also encodes the semantics of its surroundings. The upper and lower context encoding mechanisms can be described as follows:
\begin{equation}
T_i^{\text{context}} = T_i + \text{ContextEncode}\left( \{ h_j \mid j \in \mathcal{C}(i) \} \right),
\end{equation}
where \( \mathcal{C}(i) \) represents the upper and lower context set of node \( i \), and \( \text{ContextEncode} \) encodes the contextual information of surrounding nodes.

Context plays a crucial role in understanding words and phrases in sequential text. For example, "fellowship" can mean friendship or scholarship, depending on the context. In Graph2token conversion, this contextual relationship corresponds to the target node's subgraphs. Graph-BERT \cite{zhang2020graph} trains a target node's embedding using sampled subgraphs from the local context, instead of the entire input graph. This enables Graph-BERT to learn effectively in standalone mode and be transferred to other applications for fine-tuning, facilitating context migration.

Sequential text has a clear structure, while graphs vary in structure and dimensionality. To achieve effective Graph2token matching, OpenGraph \cite{xia2024opengraph} introduces a unified graph tokenizer. It employs singular value decomposition and smoothing higher-order adjacency matrices to extract node core information, converting it into a low-dimensional representation. Edge information is merged into this representation, disregarding graph differences, ensuring maximum compatibility and cross-graph adaptability of the tokens. Therefore, OpenGraph can be applied to various graph datasets for node embedding without retraining.

In heterogeneous graphs, context limitations are particularly evident. Treating a single graph as a corpus can lead to poor model generalization due to its limited context compared to large-scale linguistic data \cite{9709096}. Heterogeneous graphs introduce edge attribute categories, further narrowing the context between graphs in different scenarios. HiGPT \cite{tang2024higpt} addresses this by using parametric projection to transform nodes and edges into a unified embedding space. This captures not only direct connections but also broader relational contexts, enabling the model to learn the role of specific structures under different semantics and ensuring cross-contextual transferability.

Text corpora possess strong expressive power and generalization, often conveying different contexts with minimal text. In graphs, multiple graphs may be required to represent different contexts. UniGraph \cite{he2024unigraph} introduces Text Attribute Graphs (TAGs) to encode node textual attributes as Token sequences, followed by generating initial node embeddings using LLMs. To mitigate the issue of generalization discrepancies across different graphs, UniGraph employs a random masking technique, where portions of the node's textual information are randomly obscured. This approach enhances the model's generalization capability across diverse graphs and facilitates an indirect understanding of contextual information in varying environments. This enables UniGraph to adapt to various contexts, especially those with limited contextual samples.

\subsection{Pairwise nodes2token}
Pairwise nodes2token is a method that converts pairs of nodes in graph data into token representations, effectively capturing the relational information between them. By encoding node pairs as tokens, the method retains both local and global structural information, enabling a more comprehensive analysis of the graph \cite{guo2023linkless}. Pairwise node tokenization is particularly suitable for graphs where edges represent various types of relationships, such as positive and negative correlations. This approach allows for the nuanced encoding of different relationship types into token representations. For example, in a financial transaction network, edges may represent currency transactions and credit relationships, both of which can have positive or negative implications \cite{ling2023motif}.



\subsubsection{Position}
\ 
\newline
Pairwise nodes2token encodes node pairs, considering not only the characteristics of individual nodes but also incorporating the relative position and upper and lower contextual relationships of the two nodes into the encoding process. This creates a bidirectional representation. This design enables the model to better understand the interactions and positional information between nodes. It also captures both short- and long-distance dependencies between nodes in the pairwise node token encoding. The encoding process can be expressed as follows:
\begin{equation}
T_{ij}^{\text{pos}} = T_{ij} + \text{PosEncode}(d(i,j)),
\end{equation}
where \( T_{ij} \) represents the pairwise token of nodes \( i \) and \( j \), and \( d(i,j) \) is the distance between the two nodes.

In sequence text, an important subword often corresponds to nodes in a graph that exhibit multi-faceted link characteristics. For instance, "Alice has millions of followers" implies that the graph requires numerous nodes and edges to represent Alice's significance within global information from various perspectives. Capturing the positional information of such nodes helps the model understand their semantic importance. To retain the local topological features and connectivity patterns of the target nodes in the Graph2token conversion, NTFormer \cite{chen2024ntformer} integrates multiple perspectives based on neighborhood and node attributes (topological and attribute-based) to generate a comprehensive token sequence. This approach captures multi-dimensional features of nodes from different perspectives and preserves the topological information of the graph structure, thereby achieving positional information completion.

\subsubsection{Hierarchy}
\ 
\newline
Pairwise nodes2token encodes node pairs based on different distances, capturing both short- and long-distance dependencies. It aggregates the information of node pairs with varying hop counts to represent the hierarchical relationship between the pairs. The hierarchical aggregation process can be expressed as:
\begin{equation}
T_{ij}^{(l)} = \text{concat} \left[ \left\{ T_{kl}^{(l-1)} \mid (k, l) \in \mathcal{P}(i, j) \right\} \right],
\end{equation}
where \( \mathcal{P}(i,j) \) represents the set of node pairs containing nodes \( i \) and \( j \).

Sequence text contains various short, medium, and long-distance relationships. A long-distance relationship, such as between "I" and "big tree" in the sentence "I often sit under a big tree that has grown for many years," is semantically connected despite being far apart in the explicit sequence. Traditional GNNs, when modeling the conversion of graphs to textual representation, can only capture a single relationship between two nodes. To capture semantic relationships at different distances and proximities simultaneously, N-GCN \cite{abu2020n} extracts node relationships through random wandering with varying step lengths, treating these relationships as Token inputs. This approach directly extracts specific node-pair relationships at different step lengths. N-GCN serves as a general framework applicable to various graph convolutional networks for converting graph structures into tokens.

For scene graph modeling requiring dynamic changes, time dependency can be considered a type of hierarchy. When dealing with such time series graphs, incorporating temporal changes within the Graph2token conversion is crucial. T-PAIR \cite{akujuobi2020t} represents node pairs as feature vectors constructs time series graphs, and aggregates information from nodes and their neighbors using the pre-trained GraphSAGE. This captures the temporal evolution of node pair relationships and effectively predicts future node pair relationships. T-PAIR demonstrates the potential of sequence models to handle dynamic time series graphs.

Traditional graph-based methods like GNNs can only model relationships based on existing edges. PLNLP \cite{wang2021pairwise} incorporates node-level and edge-level neighborhood encoders based on a general GNN to capture negative sample pairs with opposite semantics. By capturing positive sample pairs with actual edges in the input graph and using a negative sampler to extract negative samples, training pairs are formed. This pairwise learning approach ranks node pairs, preserving individual node features while capturing their interactions in the graph structure, i.e., their differing semantics. As a result, the token sequence reflects both the key links and the sparsity of the graph structure.

In the conversion of graphs to textual representation, attention should be paid to different types of pairwise interactions between texts. These different types of pairwise interactions can be understood in terms of a subject-verb-object structure. For example, in the sentence "I love dogs," the link between "I" and "love" and the link between "love" and "dogs" are both one-hop connections, but they represent two distinct types of relationships. Graph structure data often consider undirected and homogeneous graphs, leading existing models to focus on a single relationship between node pairs. To capture multiple types of node pair relationships, LPFormer \cite{shomer2023adaptive} introduces a new attention mechanism that dynamically learns the coded representation of node pairs for each target link. This captures multiple link predictors and converts node pairs into token sequences, adapting to the specific needs of each link and enabling sequence models to handle large-scale, complex graph structures more efficiently.

In text representation, words at different distances within a sequence can reflect various types and strengths of relationships. For example, in a simple sentence with a subject-verb-object structure, the subject is directly connected to the verb with a strong relationship. The subject and the object are separated by the verb in the sequence, representing a long-distance secondary relationship in terms of explicit word distance. However, semantically, the subject acts on the object through the verb, and this connection is equally important, forming complete semantic information. This is reflected in the graph as structural information of one-hop, two-hop, or even ten-hop from the central node. To enable GNN to capture this relationship during the conversion of graphs to textual representation, NAGphormer \cite{chen2022nagphormer} proposes a Hop2Token module that can aggregate features from multi-hop neighborhoods. This module treats each node as a series of tokens at different hop counts, explicitly encoding the long-distance relationships in the graph. It enables the complete preservation of semantic relationships at different hop counts in the graph during the conversion from graph structure to text embedding vectors.


\subsection{Group-aware nodes2token}
Group-aware nodes2token refers to the process of converting subgraphs, or groups of nodes, within graph data into serialized tokens, aiming to capture higher-order interactions and community structures to provide a more comprehensive data representation. This method is particularly suitable for graphs with clear community structures or group characteristics, such as social media community graphs and academic citation networks \cite{xu2024openp5,jhajj2024educational}. Converting group node data into tokens better reflects the community structure within the graph, making nodes within the same community more closely associated through token representation and providing richer contextual information for LLMs to better understand and process graph data. This enhances the ability to perform prediction and inference tasks in graph-based data analysis. 


\subsubsection{Hierarchy}
\ 
\newline
Group-aware nodes2token encodes nodes within the same community or group into a token, extracting multi-level semantic information from critical nodes across different levels in the path example. Representing nodes at the group level as a more compact token also enriches the contextual information of the token. This method can use a graph attention mechanism to capture the hierarchical structure within the group of nodes. It is expressed as:
\begin{equation}
T_G = \text{GroupAttention}(h_{\text{group}}, \{h_i \mid i \in G\}),
\end{equation}
where \( T_G \) is the token of the group \( G \), and \( \text{GroupAttention} \) is a mechanism that focuses on the node relationships within the group to capture its internal hierarchical structure.

In the transformation from graph to text representation, there are correspondences between node pairs in the graph and relationship pairs in the text, together forming a complete context. However, using GNNs for the conversion of graphs to textual representation cannot directly model these important nodes or capture the modifying relationships between surrounding nodes and the anchor nodes. To address this limitation, AGFormer \cite{jiang2023agformer} introduces an anchor-based Transformer architecture, utilizing Anchor-to-Anchor Self-Attention (AASA) and Anchor-to-Node Cross-Attention (ANCA) mechanisms to tokenize node pairs. The advantage of this approach is that, regardless of the input order of the token sequence, the model can still identify the significance of the anchor nodes, ensuring consistency in the output.

PHGT \cite{luheterogeneous} introduces three different types of tokens: node tokens, semantic tokens, and global tokens. Traditional node tokens are primarily used to capture interactions between nodes within a local receptive field. PHGT extracts a series of meta path instances from the subgraph surrounding each target node, with each instance treated as a semantic token. The construction of these meta paths is based on different node and edge types in the heterogeneous graph, such as distinguishing between subject-verb structures or verb-object structures to capture heterogeneous semantics. The primary global token represents a cluster of structurally and semantically related nodes in the graph, which allows for the capture of long-range semantic-aware information in heterogeneous graphs. 

In sequence text, certain words and subwords are tightly bound together to express a single meaning. For example, in the context of molecular graphs, a benzene ring is an indivisible structure, and breaking it down would lead to semantic confusion. Therefore, during the conversion to textual representation, it should be treated as a subgraph token. Based on this context, SimSGT \cite{liu2024rethinking} introduces a GNN-based tokenizer that decomposes the molecular graph into subgraphs, which are then processed as tokens. The model utilizes a graph autoencoder and a re-masking strategy to decode the original labels of the graph, enabling a better understanding of the complex interactions within molecules. During training, random masking hides portions of the graph to perform inference decoding, training the model to effectively use available information even when key nodes are inaccessible, thereby identifying subgraph regions that should be considered as a whole. 

Heterogeneous graphs, which contain various types of edges and nodes, introduce more semantic information but also greatly increase the difficulty of graph processing. GTN \cite{yun2019graph} treats nodes and relationships in heterogeneous graphs as learnable meta-paths, transforming local substructures in the graph into tokens. By leveraging the attention mechanism of graph Transformer layers, GTN learns the relationships between different types of edges. This meta-path to token transformation method also provides a certain level of interpretability, as it explicitly shows which nodes and edge types are combined and connected, effectively capturing the local structural information in heterogeneous graphs.

\subsubsection{Context}
\ 
\newline
When addressing the context problem, the Group-aware nodes2token method encodes the global context to express the complex structure and semantics within the graph. It retains the local semantic information of the nodes in the group, preserving the interactions between group members and forming a complete context. The global context encoding process can be expressed as:
\begin{equation}
T_G^{\text{context}} = T_G + \text{GlobalContextEncode}(\{h_j \mid j \in G\}),
\end{equation}
where \( \text{GlobalContextEncode} \) represents the function that encodes the global context information of the group.

Text sequences have rich and large corpora, often nesting contexts to express information. To more effectively express this kind of nested context in the transformation from graph to text representation, it can be aligned with subgraph nesting in the graph. GIG \cite{wang2024graph} proposes an innovative Graph-in-Graph Neural Network, where each vertex in the graph is represented as a subgraph token. Two modules, i.e., vertex-level update and global-level update, are introduced to update local vertex information while enabling global context-based information exchange and updates across all vertices. This enables the capture of graph tokens influenced by multiple contextual layers. 

When feeding processed graph tokens into LLMs, the questions asked may not necessarily require all the graph structure, meaning not all the corpus information will be used. To precisely utilize the corresponding corpus resources, G-retriever \cite{he2024g} proposes a similarity-based subgraph retrieval method. First, pre-trained LLMs are used to decompose the relationships and attributes in the graph into a series of tokens. Then, a retrieval-augmented generation strategy is introduced, where the most relevant nodes and edges in the graph structure are extracted based on a given query (e.g., "What is the name of Justin Bieber's brother?") to form a compact subgraph, achieving a corpus match between the prompt semantic information and the most relevant graph structural information.

To capture the contextual relationships of each node within a graph structure, GraphPrompter \cite{liu2024can} extracts k-hop subgraphs for each node and encodes them into structural embeddings using a GNN. Subsequently, a projection layer is employed to project these embeddings into a vector space compatible with LLMs. By combining the node embeddings with text embeddings using a text embedder, the resulting embeddings are fed into LLMs as soft prompts, effectively integrating graph structural information with textual semantic information. This enables the capture of contextual semantic information.

\subsection{Holistic nodes2token}
Holistic nodes2token methods convert entire graphs into sequences of tokens, capturing global structures and overall relationships, thereby providing a comprehensive graph representation suitable for LLMs. This approach is particularly advantageous for tasks requiring a holistic view of the graph, such as graph classification and the analysis of molecular graphs, where understanding the complete structure and interdependencies is crucial. By transforming the entire graph into a token sequence, these methods enable LLMs to receive and utilize these representations, thereby leveraging global patterns and interactions to enhance the accuracy and depth of graph-related tasks.




\subsubsection{Alignment}
\ 
\newline
The holistic nodes2token tokenization process captures the relative positions and connections between nodes within an entire graph structure, allowing for a comprehensive representation of the graph and alignment with textual information for LLMs to process. To embed the entire graph into the token space of an LLM, a global graph embedding technique is applied, aggregating and projecting the features of all nodes into the token space. This can be expressed as:
\begin{equation}
T_{\text{all}} = \text{GraphEncode}(\{h_i \mid i \in G\}) + \text{Align}_{\text{LLM}}(T_{\text{all}}^{\text{text}}),
\end{equation}
where \( T_{\text{all}} \) represents the global token of the graph, and \( \text{GraphEncode} \) aggregates the features of all nodes, while \( \text{Align}_{\text{LLM}} \) aligns it with the text token space.

Domain-specific graph structures carry valuable additional information. To effectively incorporate this information into textual representations and input it into large models, aligning external information from different modalities with the graph structure's internal information is crucial. This enhances the overall graph information by integrating diverse modalities. Git-mol \cite{liu2024git} employs GNN encoding for molecular structures, tokenizing subgraphs as high-dimensional vectors. These vectors are aligned into a unified potential space with multimodal data (graph, image, and text) using the GIT-Former module. The unique cross-modal attention mechanism establishes precise connections between graph, image, and text, ensuring that each modality's features are fully reflected in the final molecular representation. This demonstrates the potential of sequence models to handle multimodal data.

Traditional approaches treat molecular structure diagrams and text descriptions as separate information sources. To utilize them in LLMs, a modal alignment process is essential, MoMu \cite{su2022molecular} encodes molecular diagrams as tokens using Graph Isomorphism Networks (GINs) and combines them with the BERT model for natural language encoding. Contrast learning establishes the link between molecular structure and language description, maximizing the similarity of relevant graph-text pairs and encoding them into a joint representation space.

To improve the generalization of graph tokens, some studies propose instruction fine-tuning. Instructmol \cite{cao2023instructmol} introduces a two-phase approach: instruction alignment and training followed by instruction fine-tuning. It uses a frozen graph encoder to transform molecular graphs into tokens and project them into an LLM-suitable embedding space. The weights of LLMs and the aligned projector are then updated based on task-specific instructions. This effectively aligns molecular graph structures with text tokens and enhances model generalization across different contexts.

Traditional discrete tokens are limited by fixed vocabularies and cannot express continuous information. To tackle this problem, GraphToken \cite{perozzi2024let} designs an efficient encoding method to convert graph structure data into soft tokens suitable for LLMs. Unlike fixed vocabulary-limited discrete tokens, soft tokens are continuous vector representations optimized through training, thus better expressing graph structure features. GraphToken directly combines graph structure information with textual cue alignment while reducing computational requirements.

Graph2Token \cite{wang2024graph2token} uses GNN pairs to encode molecular graphs into graph embedding vectors, which are mapped to an LLM embedding space through a lightweight projection layer. The pre-trained vocabulary of LLMs is used to re-represent graph Tokens, aligning them with text Tokens in the same semantic space through a trainable alignment function. This enables Graph2Token to transform graph structures into LLM-adaptable Token forms without complex fine-tuning.

MolCA \cite{liu2023molca} uses a cross-modal projector to map 2D structural information of molecular graphs into the text space of LLMs. A Q-Former module converts the graph encoder's molecular graph representation output into a 'soft prompt' understandable by LLMs, enabling graph-text representation alignment. Multi-targeting and different attentional masking strategies achieve effective graph-text interaction.

The GraphLlava \cite{argatu2024joint} architecture seamlessly integrates graph and language models by combining graph and text data. Graph data is encoded through the GRIT module, while text data is processed through the standard Transformer model. The embedded representations of both are weighted and combined to generate the final output through the predictive part of LLMs. This approach enables cross-modal information alignment, fully utilizing both graph structural and textual semantic information.

\subsubsection{Position}
\ 
\newline
The Holistic nodes2token tokenization converts the graph into a sequence of node and edge tokens, allowing the sequence model to retain the positional information of nodes from multiple paths. This process can be abstracted as leveraging the distance matrix of the entire graph for encoding. It is expressed as:
\begin{equation}
T_{\text{all}}^{\text{pos}} = T_{\text{all}} + \text{PosEncode}(D),
\end{equation}
where \( D \) is the distance matrix of the entire graph, and \( \text{PosEncode}(D) \) encodes the relative positions of nodes in the global graph.

In real-world tasks, graph data often suffers from random nodes or has missing edges, such as when sensitive data is inaccessible. Traditional GNN may produce biased node representations due to incomplete feature propagation, leading to the loss of positional and semantic information. To address this issue, GraphGPT \cite{zhao2023graphgpt} leverages (semi-)Eulerian paths to transform graphs into token sequences representing nodes, edges, and attributes. Nodes are renumbered according to the path order, resulting in a sequence containing tokens for nodes, edges, and attributes. As multiple Eulerian paths can exist for a graph, GraphGPT can generate diverse sequences by randomly selecting different paths, enabling data augmentation. This allows sequence models to preserve positional information from multiple perspectives and facilitate task-specific enhancements.
In addition, GraphToken transforms graph-structured data into soft tokens, which are continuous vector representations that preserve positional information.


\section{Selecting the Proper Large Language Model for Graphs}
\subsection{Graph Characteristics Analysis}
When selecting appropriate LLMs for practical graph-related tasks, it is crucial to consider the specific properties and types of the relevant graphs \cite{hamilton2017inductive}. We categorize graph properties into AMR graphs, KG, attributed graphs, heterogeneous graphs, community graphs, and spatial graphs, aiming to provide subsequent researchers with an initial understanding of model selection based on specific graph data.




\subsubsection{Textual Graphs}
\
\newline
Textual Graphs are specialized graphs where nodes and edges represent textual elements or relationships derived from text, such as Citation networks, knowledge graphs, or AMR graphs. These graphs are characterized by their ability to encode semantic relationships between text fragments, making them crucial for natural language processing tasks. 

For Textual Graphs, models like Graph2text such as Gpt4graph \cite{guo2023gpt4graph} and LLM4DyG \cite{zhang2023llm4dyg} are particularly suitable. These models are designed to translate the structural information embedded in the graphs into coherent textual representations. By converting graph structures into text, Graph2text models enable LLMs to process and generate insights from the underlying textual relationships more effectively. This approach significantly improves the understanding and generation of text-based knowledge, making it invaluable for tasks like information retrieval and content generation.

\subsubsection{Attributed Graphs}
\ 
\newline
Attributed graphs are characterized by nodes and edges that contain additional data or attributes, such as labels, weights, or other properties. These graphs are commonly found in social networks and knowledge graphs, where the extra information is crucial for understanding the underlying relationships. 


For these types of graphs, Node2token models such as OpenGraph \cite{xia2024opengraph} and VQGraph \cite{yang2023vqgraph} are particularly suitable. These models effectively handle the complexity by preserving attributes during tokenization. In this way, they enable LLMs to process and analyze the enriched data more accurately, thereby improving the overall understanding and extraction of meaningful insights from the graph.
\subsubsection{Heterogeneous Graphs}
\ 
\newline
Heterogeneous graphs, also known as heterogeneous information networks, are characterized by the presence of multiple types of nodes and edges, each representing different entities and relationships within a complex system. These graphs are common in various domains, including social networks, knowledge graphs, and multi-relational databases, where the diversity of node types and edge types reflects the rich and varied interactions among entities.

Pairwise nodes2token methods are particularly well-suited for representing heterogeneous graphs. These methods convert pairs of nodes into token representations, effectively capturing the relational information between them. This is crucial for tasks such as link prediction and relation classification, where understanding the interactions between node pairs is essential. By encoding node pairs as tokens, these methods preserve both local and global structural information, enabling a comprehensive analysis of the graph. For example, in a financial transaction network, edges may represent different kinds of transactions and credit relationships. This precise encoding enables LLMs to process and analyze the data more effectively, enhancing the accuracy and depth of graph-related tasks.
\subsubsection{Community graphs}
\ 
\newline
Community graphs are distinguished by their structure, consisting of densely connected groups of nodes that form clusters. These graphs are prevalent in social networks and biological networks, where the community structure plays a significant role in the dynamics and interactions within the network. For community graphs, Group-aware nodes2token models like SimSGT \cite{liu2024rethinking} and GraphLLM \cite{chai2023graphllm} are particularly effective. These models are designed to tokenize community structures while preserving the clustering information essential for analysis. By capturing the inherent community structures, these models provide an input format that allows LLMs to process and understand the data more effectively. This approach enhances the ability of LLMs to analyze and interpret complex community dynamics, leading to more accurate and insightful results.

\subsection{Optimized Implementation of LLM4graph}
\label{trick}
Different LLMs may exhibit limitations in their understanding of graph structures when faced with different types of graph task. Adopting targeted optimization strategies can improve the performance of LLMs in graph tasks. This section introduces prompting and fine-tuning strategies that have been proven to be effective means of enhancing the graph reasoning capabilities of LLMs, according to their respective technical types.

\subsubsection{Prompt}
\
\newline
Prompting is a technique that supplies the model with contextual information such as a question or task prompt to guide it toward generating specific responses or performing defined actions. Based on how and why prompts are constructed, we categorize prompts into three groups: hard prompts, soft prompts, and prompt engineering approaches.

\textbf{Hard prompts} are artificially designed and are usually fixed in the form of natural language, explicitly outlining the steps and requirements of the task. Hard prompts have been observed in systems such as NLGraph \cite{wang2024can}, LLM4DyG \cite{zhang2023llm4dyg}, and GIMLET \cite{zhao2023gimlet}, with examples as follows:
\bigskip 
\begin{mdframed}
\textbf{Build-a-Graph Prompting:} "Given<graph description>. Let's construct a graph with the nodes and edges first."
\end{mdframed}
\bigskip 
\begin{mdframed}
\textbf{Algorithmic Prompting:} "The basic idea is to start at one of the nodes and use DFS to explore all of its adjacent nodes. At each node, you can keep track of the distance it takes to reach that node from the starting node. Once you have explored all the adjacent nodes, you can backtrack and pick the node that has the shortest distance to reach the destination node."
\end{mdframed}
\bigskip 
\hspace{10pt} \textbf{Soft prompts} are automatically optimized through training and learning, often being more flexible and adaptable. These prompts are generated based on embeddings and do not require explicit task descriptions, allowing them to dynamically adjust according to model training. Soft prompts have been observed in systems such as GPT4Graph \cite{guo2023gpt4graph}, Graph-ToolFormer \cite{zhang2023graph}, LLaGA \cite{chen2024llaga}, HiGPT \cite{tang2024higpt}, UniGraph \cite{he2024unigraph}, GraphEdit \cite{guo2024graphedit}, and Graph2Token \cite{wang2024graph2token}, with examples as follows:
\bigskip 
\begin{mdframed}
\textbf{LLaGA \cite{chen2024llaga}:} "A chat between a curious user and an artificial intelligence assistant. The assistant gives helpful, detailed, and polite answers to the user's questions. USER: Given you a node: <node sequence>...Please tell me..."
\end{mdframed}
\bigskip 
\begin{mdframed}
\textbf{Graph2Token \cite{wang2024graph2token}:} "The HOMO-LUMO gap is the energy difference between the highest occupied molecule orbital (HOMO) and the lowest unoccupied molecule orbital (LUMO) in a molecule. What is the HOMO-LUMO gap of this molecule? <graph>"
\end{mdframed}
\bigskip 
\hspace{10pt}  \textbf{Prompting techniques} encompass methods such as Chain of Thought (CoT) \cite{wei2022chain} and few-shot learning that enhance the model's reasoning capabilities. The CoT prompting technique has been utilized in GraphGPT.
\bigskip 
\begin{mdframed}
\textbf{CoT \cite{wei2022chain}:} "Please think about the categorization in a step-by-step manner."
\end{mdframed}
\bigskip 

\subsubsection{Fine-tuning}
\
\newline
Fine-tuning adjusts an LLM's parameters (or adds specialized encoders) to more effectively incorporate graph information. Depending on the approach used for integrating graph information into the LLM, we can classify fine-tuning into three types: prefix fine-tuning, similarity fine-tuning, and corpus fine-tuning.

\textbf{Prefix Fine-tuning} embeds essential task information at the beginning of the model's input, enabling rapid adjustments to its generation strategy. Prefix fine-tuning is commonly applied in tasks involving specific graph structures, such as in GraphLLM and GraphLlava, where linear projections or bias terms of the graph are embedded in the prefix to help the model capture the structural information more effectively. This approach is particularly useful for tasks when the model must directly access contextual information, ultimately improving the accuracy of task processing.

\textbf{Similarity Fine-tuning} leverages the similarity between data instances or between generated text and target text to improve how LLMs handle node heterogeneity and generate accurate responses. By training the model to capture the similarity between labels, the model can better identify homogeneity in the graph structure or deal with noise issues. For example, GraphQA scores the alignment between the generated and standard answers, adjusting the combination of the transformation function for selecting edges and nodes. Retrieve-Rewrite-Answer fine-tunes LLMs to generate rewritten text that is more in line with the KG structure by maximizing the similarity between the target text related to the KG structure and the generated text output based on the triple-form text obtained by sampling by LLMs. GraphEdit \cite{guo2024graphedit} uses similar paper titles and abstracts to enhance the model's ability to uncover hidden patterns in the graph. An example of the instructions used by GraphEdit \cite{guo2024graphedit} for fine-tuning LLMs is as follows:
\bigskip 
\begin{mdframed}
\textbf{Graphedit \cite{guo2024graphedit}: }Based on the title and abstract of the two paper nodes. Do they belong to the same category among \{Category\_0\}, \{Category\_1\}, \{Category\_2\}, \ldots ? If the answer is "True", answer "True" and the category, otherwise answer "False". The first paper: \{Title\}, \{Abstract\}. The second paper: \{Title\}, \{Abstract\}. A: \{True or False\}, \{Category\}.
\end{mdframed}
\bigskip

\textbf{Corpus Fine-tuning} employs multi-modal corpora to align the model for downstream graph task, ensuring effective cross-modal intergration. For instance, Graph-ToolFormer \cite{zhang2023graph} fine-tunes LLMs using a ChatGPT-enhanced graph reasoning prompt dataset, enabling the model select the appropriate external APIs and insert the prompt at the right stage. InstructMol aligns molecule-text pairs through two-stage fine-tuning process, allowing LLMs to fully capture molecular graph structures. Meanwhile, GIT-Mol leverages contrastive learning to align graph, text, and image modalities, thereby enhancing both stability and integration across diverse domains.

\begin{table}[htbp]
\centering
\caption{\textbf{LLM-related Resources for Graph Tasks}. The effects of each method corresponding to the trick are (a) Enhancing graph structure understanding, (b) Improving task versatility, (c) Saving computing resources, and (d) Enhancing the interpretability of answers. `-' means the original paper does not provide specific details.}
\label{sources}
\vspace{-0.3cm}  
\renewcommand{\arraystretch}{0.9}  
\tiny  
\begin{adjustbox}{max width=\textwidth}  
\begin{tabular}{ccccccc}
\toprule
\textbf{Method} & \textbf{LLM} & \textbf{Trick} & \textbf{Effect} & \textbf{Hardware Resource} & \textbf{Link} \\
\midrule
\multirow{2}{*}{GraphQA \cite{fatemi2023talk}} & \multirow{2}{*}{Palm 2} & Hard prompt & (b) & \multirow{2}{*}{TPU v4 4$\times$4} & \multirow{2}{*}{-} \\
\cmidrule(lr){3-4}
 &  & Similarity fine-tuning & (d) &  & \\
\cmidrule(lr){1-6}
\multirow{2}{*}{NLGraph \cite{wang2024can}} & \multirow{2}{*}{TEXT-DAVINCI-003} & Hard prompt & (b) & \multirow{2}{*}{-} & \multirow{2}{*}{\href{https://github.com/arthur-heng/nlgraph}{code}} \\
\cmidrule(lr){3-4}
 &  & Hard prompt & (a) &  & \\
\cmidrule(lr){1-6}
LLM4DyG \cite{zhang2023llm4dyg} & GPT-3.5 & Hard prompt & (a) & - & \href{https://github.com/wondergo2017/llm4dyg}{code} \\
\cmidrule(lr){1-6}
Gpt4graph \cite{guo2023gpt4graph} & InstructGPT-3 & Soft prompt & (a) & - & - \\
\cmidrule(lr){1-6}
\multirow{2}{*}{Graph-ToolFormer \cite{zhang2023graph}} & ChatGPT & Hard prompt & (a) & NVIDIA Ampere A100 80 GB × 1 & \multirow{2}{*}{\href{https://github.com/jwzhanggy/Graph_Toolformer}{code}} \\
\cmidrule(lr){2-4}
 & GPT-J & Corpus fine-tuning & (b) & NVIDIA GeForce RTX 4090 × 1 & \\
\cmidrule(lr){1-6}
\multirow{2}{*}{Retrieve-Rewrite-Answer \cite{wu2023retrieve}} & Llama-2/Flan-T5 & Similarity fine-tuning & (a) & \multirow{2}{*}{NVIDIA Tesla V100 $\times$ 4} & \multirow{2}{*}{\href{https://github.com/wuyike2000/retrieve-rewrite-answer}{code}} \\ 
\cmidrule(lr){2-4}
 & Llama-2/T5/T0/ChatGPT & Hard prompt & (b) & & \\
\cmidrule(lr){1-6}
GraphEdit \cite{guo2024graphedit} & Vicuna-v1.5 & Similarity fine-tuning & (a) & - & \href{https://github.com/HKUDS/GraphEdit}{code} \\
\cmidrule(lr){1-6}
GIMLET \cite{zhao2023gimlet} & GPT-3.5 & Hard prompt & (b) & - & - \\
\cmidrule(lr){1-6}
LLaGA \cite{chen2024llaga} & Vicuna-7B & Soft prompt & (a) & - & \href{https://github.com/zhao-ht/GIMLET}{code} \\
\cmidrule(lr){1-6}
HiGPT \cite{tang2024higpt} & GPT-3.5 & Soft prompt & (a) & - & \href{https://github.com/HKUDS/HiGPT}{code} \\
\cmidrule(lr){1-6}
GraphGPT \cite{tang2024graphgpt} & GPT-3.5 & Prompt technology & (d) & NVIDIA A100 40G $\times$ 4 & \href{https://github.com/HKUDS/GraphGPT}{code} \\
\cmidrule(lr){1-6}
GraphTranslator \cite{zhang2024graphtranslator} & GPT-3.5 & Hard prompt & (a) & NVIDIA Tesla V100 32G $\times$ 4 & \href{https://github.com/alibaba/GraphTranslator}{code} \\
\cmidrule(lr){1-6}
UniGraph \cite{he2024unigraph} & ChatGPT & Hard prompt & (b) & NVIDIA A100 40G $\times$ 8 & - \\
\cmidrule(lr){1-6}
GIT-Mol \cite{liu2024git} & GPT series & Prefix fine-tuning & (a) & - & - \\
\cmidrule(lr){1-6}
\multirow{2}{*}{GraphLLM \cite{chai2023graphllm}} & \multirow{2}{*}{GPT-3.5/GPT-4, LLaMA 2} & Prefix fine-tuning & (a) & \multirow{2}{*}{NVIDIA A100 80G $\times$ 4} &  \multirow{2}{*}{\href{https://github.com/mistyreed63849/Graph-LLM}{code}} \\
\cmidrule(lr){3-4}
 &  & Streamlined tokens & (c) & & \\
\cmidrule(lr){1-6}
\multirow{2}{*}{G-Retriever \cite{he2024g}} & \multirow{2}{*}{Llama2-7B} & Hard prompt & (a) & \multirow{2}{*}{NVIDIA A100 80G $\times$ 2} & \multirow{2}{*}{\href{https://github.com/XiaoxinHe/G-Retriever}{code}} \\
\cmidrule(lr){3-4}
 &  & Precise retrieval & (c) & & \\
\cmidrule(lr){1-6}
GraphPrompter \cite{su2022molecular} & LLAMA2-7B & Soft prompt & (a) & NVIDIA A100 80G $\times$ 2 & \href{https://github.com/franciscoliu/graphprompter}{code} \\
\cmidrule(lr){1-6}
InstructMol \cite{cao2023instructmol} & Vicuna-7B & Corpus fine-tuning & (a) & RTX A6000 48GB $\times$ 4 & \href{https://idea-xl.github.io/InstructMol}{code} \\
\cmidrule(lr){1-6}
Graphtoken \cite{perozzi2024let} & LLama2 & Soft prompt & (a) & - & \href{https://github.com/ZeLeBron/Graph2Token}{code} \\
\cmidrule(lr){1-6}
Graph2Token \cite{wang2024graph2token} & LLama2-7B & Soft prompt & (a) & NVIDIA GEFORCE GTX4090 $\times$ 1 & - \\
\cmidrule(lr){1-6}
GraphLlava \cite{argatu2024joint} & GPT4/TinyLlama & Corpus fine-tuning & (b) & - & - \\
\bottomrule
\end{tabular}
\end{adjustbox}
\end{table}

\subsection{Practical Guideline}
In this section, we offer a series of guidelines for selecting models based on the techniques and corresponding effects that improve LLMs performance in specific domains or graph-related tasks, as discussed in Section \ref{trick}. We also summarize the hardware requirements and open-source resources for various methods in Table \ref{sources}.

First, from the computational resource perspective: fine-tuning often requires updating parameters in open-source LLMs such as LLaMA, significantly increasing resource demands like VRAM as the model's scale grows. In contrast, using LLMs like ChatGPT via API calls only allows for prompt improvements in graph understanding without requiring internal model adjustments. This reduces hardware requirements, as no additional training is needed to alter the model's structure or weights. Table \ref{sources} outlines the hardware resources required by each method, offering researchers guidance on model selection. We also propose two tricks—streamlined tokens and precise retrieval—as potential approaches for reducing computational overhead.

Second, from the task-specific data processing perspective, fine-tuning typically requires a substantial amount of high-quality data to train the model effectively, allowing it to understand graph structures in-depth and transfer learning efficiently across different graph tasks. Therefore, when working with limited datasets or domain-specific knowledge, it may be more appropriate to use prompt-based tricks. Furthermore, if the downstream graph task involves sensitive data, such as friend relationships data in social networks, prompt-based tricks should be employed to avoid exposing private information to external models. 

However, it's important to note that prompt tricks lack transparency in the model's internal mechanisms, which can limit their effectiveness for certain graph tasks due to the inherent constraints of the LLM's ability to generalize to graph-specific tasks. Conversely, fine-tuning provides granular control over the model's behavior through code and adjustable parameters, making it more adaptable to specific tasks.
Finally, these tricks can be used in a complementary manner, depending on the application and requirements. For example, a model can be fine-tuned first, followed by using well-designed prompts to guide LLMs in generating output in the desired format.

\section{Open problems and Future Directions}
Despite significant progress in integrating graph learning with LLMs, several critical open questions remain, offering opportunities for further research and development. This section explores five open issues in the realm of LLMs for graphs and envisions future research directions.

To address these challenges, future research should focus on developing advanced models and algorithms that can effectively utilize the semantic richness of KGs. This includes creating methods for seamless integration of dynamic and large-scale data, ensuring that LLMs can adapt and maintain performance. Enhancing the ability of LLMs to exploit the detailed knowledge within KGs will significantly improve their performance in complex, knowledge-intensive applications.

\subsection{Domain-specific Instructions and General Learning Framework}
Graph2text methods mainly design methods and prompts according to a specific domain (e.g., AMR graphs and KGs).
Although they perform well on specific domains, these methods cannot freely transfer other kinds of graphs.
Thus, there is a strong need to design general instructions or prompts to tackle any kind of graph.
Specifically, these instructions should be rooted in the structures of graphs rather than the additional information of graphs (i.e., texts).
A potential solution is directly describing structures for LLMs to judge the existence or properties of nodes/edges.
For example, Guo et al. \cite{guo2024graphedit} exploited homophily property assumption to design prompts for predicting node classes. 
However, their instructions still require text descriptions of nodes, and they only test their methods on citation graphs with rich text.
We suggest that researchers develop general instructions by borrowing ideas from existing benchmarks \cite{wang2024can,wu2024grapheval2000,li2024glbench}.
Through general instructions, LLMs may have the potential to deal with graph-related tasks alone, without the help of graph-based methods.

Graph2token methods concentrate on learning general structural information as token representation, regardless of the specific domain of the graph.
Although they have a general ability to learn any type of graphs, they cannot be customized to further improve performance in a specific domain.
However, it is hard to embed prior knowledge into the encoding stage of Graph2token.
Thus, there is a strong need to develop a modular framework to combine domain-specific information with Graph2token methods.
We propose to combine Graph2text methods and Graph2token methods.
Specifically, Graph2text serves as domain-specific modules, while Graph2token serves as generally structured learning modules.
As a result, LLMs can achieve dominant performance through both domain-specific information and general learning ability.

\subsection{Theoretical Power of LLMs}
LLMs demonstrate remarkable capability across various tasks, yet their theoretical power in dealing with graph structures, especially regarding permutation invariance \cite{keriven2019universal} and geometric transformations \cite{satorras2021n}, remains a critical area of exploration. When dealing with graphs, one must consider whether LLMs should maintain consistent outputs if only the order of nodes or edges changes. This is especially relevant in tasks such as node classification and link prediction, where the sequence of inputs should not affect the outcome. Moreover, for geometric graphs and physical systems where nodes include spatial information, LLMs must exhibit equivariance under operations such as rotations, translations, and reflections \cite{joshi2023expressive}. The challenge here lies in ensuring that LLMs can process these transformations without compromising the integrity of the graph’s structural information.

To address these challenges, it is necessary to develop models or frameworks that enhance the ability of LLMs to respect these invariances. To address the issue of permutation invariance, it is proposed to integrate a subgraph-based structural encoding mechanism when converting graphs into text or tokens. This encoding does not rely on the input order of nodes or edges but is instead based on the topological features of the graph, focusing on the local neighborhood topology surrounding individual nodes. This approach mitigates the impact of input permutation changes. To recognize various geometric transformations, such as rotations and translations in geometric graphs, a potential solution is to incorporate symmetry operations during the Graph2text or Graph2token conversion process. This could involve standardizing the different transformations of the input graph by applying operations such as rotations and reflections to the embedded node representations, enabling the model to identify equivalence.

\subsection{Fair Graph Learning}

LLMs have demonstrated effectiveness in reducing bias when handling sensitive social and gender-related issues \cite{luo2024algorithmic}, ensuring that generated outputs are more politically and ethically equitable. Compared to traditional graph learning algorithms, LLMs can enhance fairness by designing appropriate prompts, without the need for complex algorithmic modifications. This is especially beneficial for fairness assessments at both individual and group levels, making LLMs a key tool for improving fairness in graph learning algorithms.

Unlike traditional data (e.g., text and image), fairness in graph data involves multiple interconnected levels. For example, individual fairness pertains to specific nodes, while group fairness focuses on the relationships and structures within the graph, which are inherently absent in traditional data formats \cite{dong2023fairness}. Additionally, bias in a single node can propagate and affect the entire network. In traditional graph learning algorithms, ensuring fairness across these levels often requires designing intricate debiasing algorithms to fully address the structural complexity of graph data and its inherent group relationships. LLMs offer a simplified approach by using targeted prompts, allowing the model to automatically analyze and address fairness concerns from specific angles in graph data. For example, LLMs can readily achieve individual fairness or group fairness by designing appropriate prompts. Consequently, LLMs can not only simplify the design of complex algorithms but also effectively meet the fairness demands across multiple levels.

\subsection{Efficiency and Scalability}
LLMs often struggle with highly imbalanced graphs, especially when nodes have \emph{very large degrees} (e.g., influential users in social networks with hundreds of friends).
Handling each neighbor individually forces the model to perform numerous forward passes, drastically increasing computational overhead for a single node. On the other hand, processing all neighbors at once leads to an exceedingly long input sequence-especially considering that each node or edge may also carry rich textual information. Because LLMs cannot effectively process extremely long contexts, they either fail to fully capture an important node's neighborhood or resort to segmentation that risks structural loss.

To address these issues, hierarchical attention and neighborhood sampling should be incorporated into Graph2text and Graph2token approaches. These methods can prioritize key relationships by aggregating the most relevant neighborhood information, thereby converting local structures into generalized sequence texts or sequence tokens. The subset of neighbors can be selected based on relevance metrics, such as the node's influence within the graph. By segmenting the neighborhood into smaller, more manageable parts, it is possible to avoid exceeding input length limits while minimizing structural loss. Each segment can be processed independently, and the outcomes are later merged into a comprehensive neighborhood view.

Moreover, the numerous number of parameters in LLMs may be redundant for processing graph-structured data since graphs are only a subdomain of real-world knowledge. To tackle only graph data, we do not need the full language power of LLMs, which is inefficient and not environmentally friendly.
A more efficient alternative is to distill or prune the model for Graph2text and Graph2token. This model would retain only the functionalities required for graph processing, creating a data dictionary that links graph structures with LLMs. Further, task-specific fine-tuning can then enhance the model, making it more efficient for graph-related tasks by focusing on relevant graph patterns and discarding unnecessary language modeling features. Depending on the specific task, relevant texts to tokens can be retrieved on demand, achieving both computational and memory efficiency.

\subsection{LLMs for Dynamic Graphs}
Dynamic graphs, which evolve in both structures and attributes over time, more accurately represent complex real-world systems such as evolving traffic networks and fluctuating e-commerce prices and user interests. Despite their practical value, integrating dynamic graphs with LLMs remains largely unexplored. One major challenge is the lack of benchmarks for dynamic graph data, making it difficult to evaluate and compare models effectively. Moreover, designing prompts that help LLMs to capture temporal evolutions in graphs is still an open challenge \cite{10026151}. Additionally, real-time updates demand algorithms that can preserve model performance amid continual changes. Future research should emphasize benchmark creation and the seamless dynamic data integration so that LLMs can adapt to these evolving structures and perform consistenly.

\section{Conclusion}
In this survey, we systematically reviewed the applications of LLMs in graph learning. We introduced a novel classification framework based on the granularity of graph data: Graph2text and Graph2token. And we identified four key challenges in converting graphs to textual representations: alignment, positioning, multi-level semantics, and context.
Guided by these challenges, We provided a comprehensive analysis of each method's technical routes, strengths, and weaknesses.
Additionally, we summarized the optimization strategies used when applying LLMs to graph tasks, such as prompting and fine-tuning, and provide an overview of the open-source availability and resource support of the models, aiming to offer practical guidance for model selection. Lastly, we highlighted five open problems and future research directions, aiming to clarify both the progress and hurdles in LLM-based graph learning and inspire further innovation in this area.

Despite their promise for handing graph-structured data, LLMs still face notable challenges-particularly in efficient graph tokenization, fairness for large and complex datasets, and integrating dynamic graph data. Moving forward, research should focus on developing more scalable and efficient tokenization methods, as well as refining LLMs to enhance their fairness and accuracy in domain-specific graph-related tasks. Such progress will open new possibilities for integrating LLMs with graphical data across a diverse range of complex domains.

\bibliographystyle{ACM-Reference-Format}
\bibliography{sample-base}


\end{document}